\documentclass{article}

\usepackage{microtype}
\usepackage{graphicx}
\usepackage{booktabs} %
\usepackage{xcolor}
\usepackage{hyperref}

\usepackage[accepted]{mlsys2022}

\usepackage{amsmath}  %
\usepackage[frozencache,cachedir=.]{minted}
\usepackage{xspace}  %
\usepackage{xurl} %
\usepackage{float}
\usepackage{enumitem} %

\usepackage{caption}
\usepackage{subcaption}  %

\newcommand{\IGNORE}[1]{}

\newcommand{\sectionshrinker}{}

\newcommand{\extrashrink}[1]{}

\newcommand{\Linput}{L_{\textrm{input}}}
\newcommand{\Lgen}{L_{\textrm{gen}}}
\newcommand{\dmodel}{d_{\textrm{model}}}
\newcommand{\dff}{d_{\textrm{ff}}}
\newcommand{\nlayers}{n_{\textrm{layers}}}
\newcommand{\nheads}{n_{\textrm{heads}}}
\newcommand{\nparams}{n_{\textrm{params}}}
\newcommand{\nchips}{n_{\textrm{chips}}}
\newcommand{\dhead}{d_{\textrm{head}}}

{\end{list}}

{\end{list}}

\mlsystitlerunning{Efficiently Scaling Transformer Inference}

\begin{document}

\twocolumn[
\mlsystitle{Efficiently Scaling Transformer Inference}

\mlsyssetsymbol{equal}{*}

\begin{mlsysauthorlist}
\mlsysauthor{Reiner Pope}{g}
\mlsysauthor{Sholto Douglas}{g}
\mlsysauthor{Aakanksha Chowdhery}{g}
\mlsysauthor{Jacob Devlin}{g}
\mlsysauthor{James Bradbury}{g}\\
\mlsysauthor{Anselm Levskaya}{g}
\mlsysauthor{Jonathan Heek}{g}
\mlsysauthor{Kefan Xiao}{g}
\mlsysauthor{Shivani Agrawal}{g}
\mlsysauthor{Jeff Dean}{g}
\vspace{-0.05in}
\end{mlsysauthorlist}

\mlsysaffiliation{g}{Google}

\mlsyscorrespondingauthor{Sholto Douglas}{sholto@google.com}
\mlsyscorrespondingauthor{Aakanksha Chowdhery}{chowdhery@google.com}

\mlsyskeywords{machine learning, LLMs, inference}

\vskip 0.3in

\begin{abstract}
We study the problem of efficient generative inference for Transformer models, in one of its most challenging settings: large deep models, with tight latency targets and long sequence lengths. Better understanding of the engineering tradeoffs for inference for large Transformer-based models is important as use cases of these models are growing rapidly throughout application areas.
We develop a simple analytical model for inference efficiency to select the best multi-dimensional partitioning techniques optimized for TPU v4 slices based on the application requirements. We combine these with a suite of low-level optimizations to achieve a new Pareto frontier on the latency and model FLOPS utilization (MFU) tradeoffs on 500B+ parameter models that outperforms the \citet{fastertransformer} suite of benchmarks. We further show that with appropriate partitioning, the lower memory requirements of multiquery attention (i.e. multiple query heads share single key/value head) enables scaling up to $32\times$ larger context lengths.
Finally, we achieve a low-batch-size latency of 29ms per token during generation (using int8 weight quantization) and a 76\% MFU during large-batch-size processing of input tokens, while supporting a long 2048-token context length on the PaLM 540B parameter model. %

\end{abstract}

]

\printAffiliationsAndNotice{}  %

\widowpenalty0
\clubpenalty0

\sectionshrinker

\section{Introduction}\label{sec:intro}

\global\csname @topnum\endcsname 0
\global\csname @botnum\endcsname 0

Scaling Transformer-based models to 100B+~\cite{brown2020language, kaplan2020scaling,rae2021scaling,chinchilla} and later 500B+ parameters~\cite{chowdhery2022palm,smith2022using} has led to state of the art results on natural language processing benchmarks. The practical utility of these large language models (LLMs) in a variety of applications makes them compelling for widespread use.
While the sequence parallelism of the Transformer architecture enables highly parallel training, efficient deployment of these models is challenging in practice because generative inference proceeds one token at a time and the computation for each token sequentially depends on the previously generated tokens. Thus, models that support efficient training at scales of thousands of chips require careful attention to parallel layout and memory optimizations to unlock the scalability needed for efficient, low-latency inference.
This paper focuses on a simple set of engineering principles that enable serving large-scale Transformer-based models efficiently in a variety of challenging production settings.

We consider the requirements of downstream applications for LLMs.
Some applications, including interactive workloads like chatbots, involve tight latency constraints \cite{thoppilan2022lamda}.
Others, including offline inference for scoring or distillation, emphasize high throughput and low cost per token at any latency.

We discuss briefly what makes generative inference of LLMs challenging.
First, large models have a large memory footprint both due to the trained model parameters as well as the transient state needed during decoding.  The model parameters generally do not fit in the memory of a single accelerator chip.  The attention key and value tensors of each layer, which we refer to as the \emph{KV cache}, must also be stored in memory for the duration of decoding.
Second, tight latency targets become especially challenging for generative inference given the much lower parallelizability of Transformer generation relative to training. The large memory footprint gives rise to a large amount of memory traffic to load the parameters and KV cache from high-bandwidth memory (HBM) into the compute cores for each step, and hence a large total memory bandwidth required to meet a given latency target. 
Finally, inference cost from the attention mechanism scales quadratically with input sequence length~\cite{sukhbaatar2019adaptive,choromanski2020rethinking, dao2022flashattention}. %

\begin{figure*}
\centering
    \begin{subfigure}{0.49\linewidth}
  \centering
  \includegraphics[height=2in]{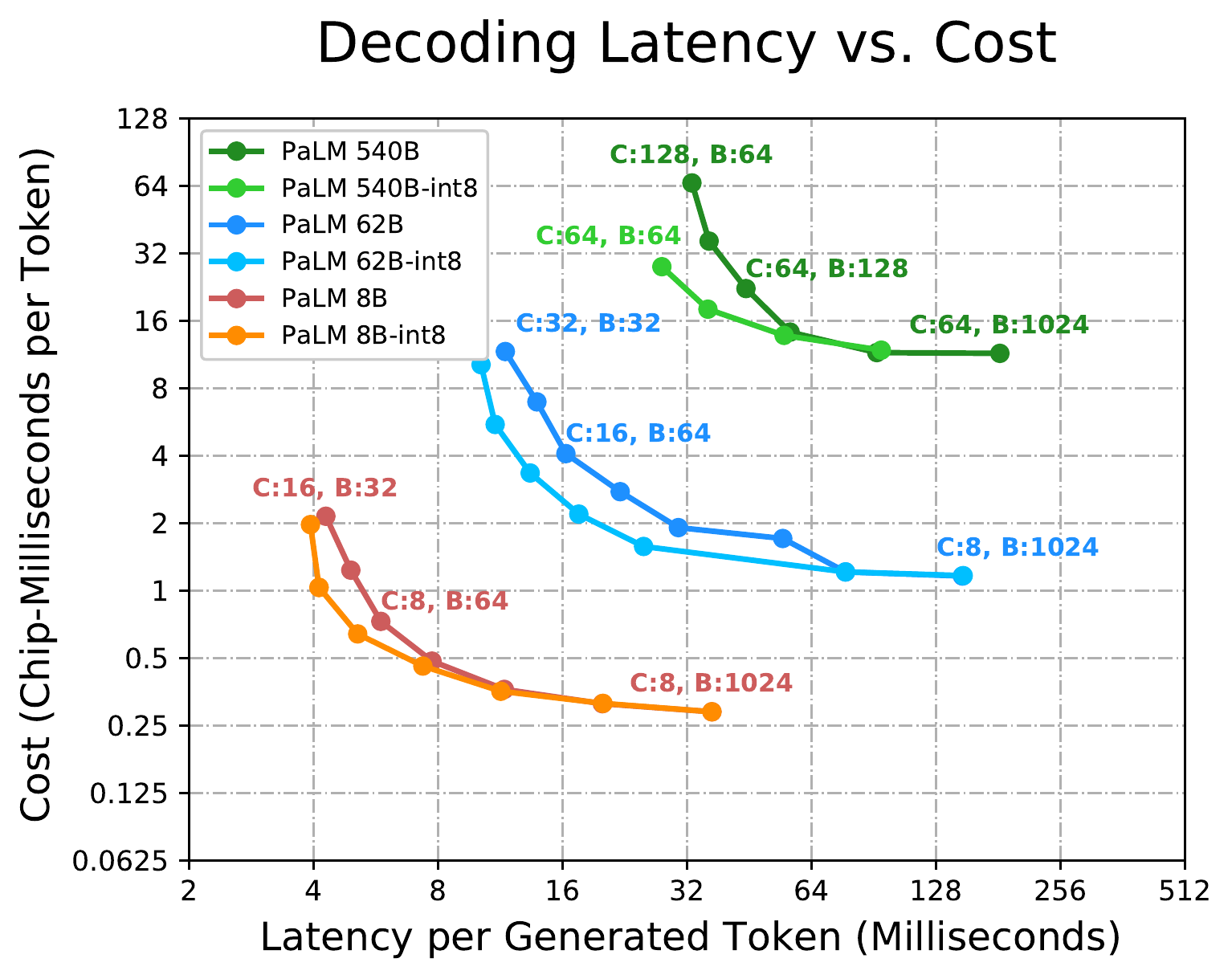}
  \label{fig:generate-cost-vs-latency}
  \end{subfigure}
\begin{subfigure}{0.49\linewidth}
  \centering
  \includegraphics[height=2in]{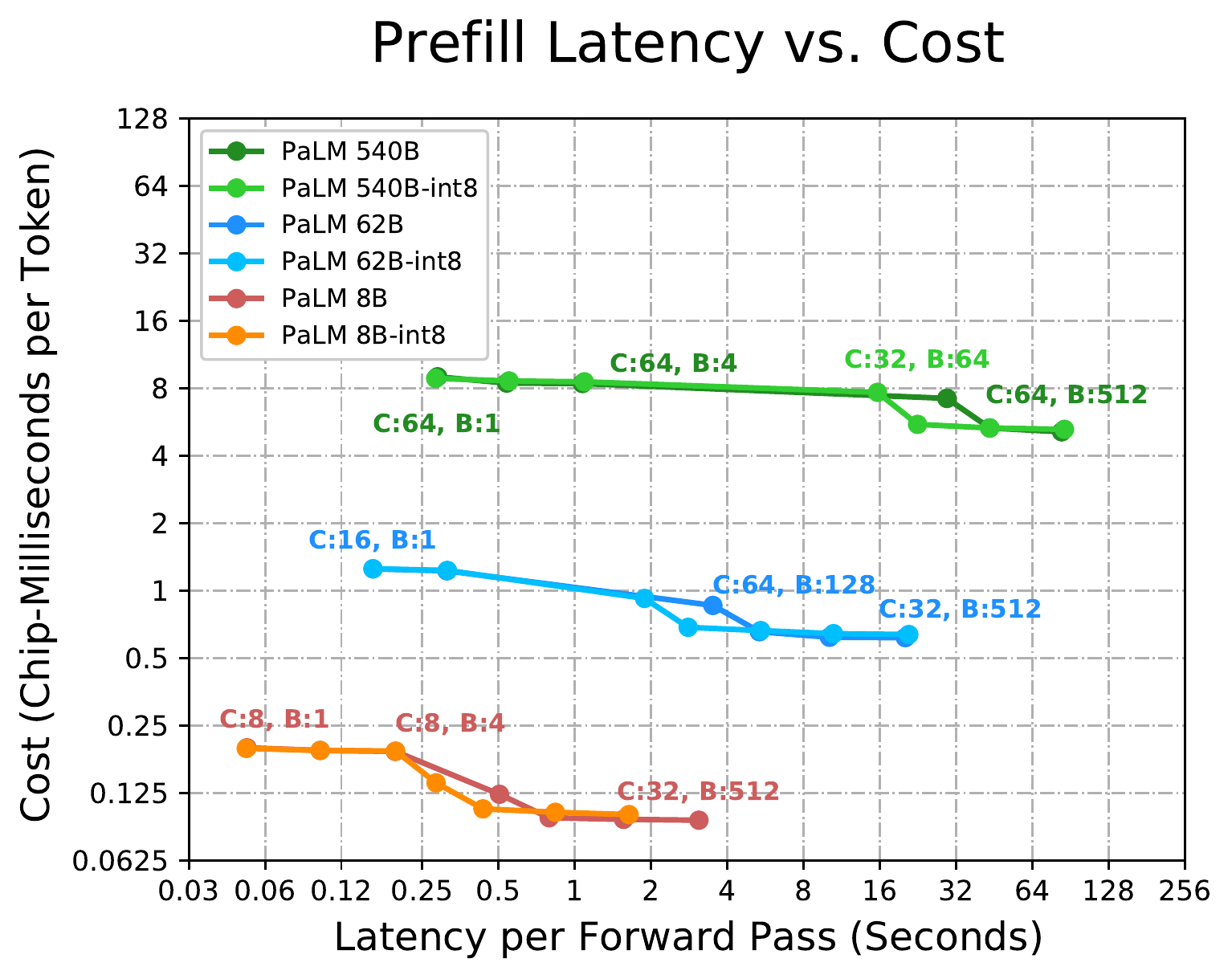}
  \label{fig:prefill-cost-vs-latency}
  \end{subfigure}
  \caption{Cost vs.\ latency for PaLM models. We use a context length of 2048. Points in each line represent the Pareto frontier of efficiency versus latency. Chip count is $C$, batch size is $B$. Left: latency per token for generating 64 tokens, assuming the context has already been processed. Right: time to process 2048 input tokens; excludes the time to generate any output tokens. Tables~\ref{table:cost-vs-latency} and~\ref{table:cost-vs-latency-62b} show details on a few specific scenarios from the Pareto frontier where the applications have low-latency or high-throughput requirements.}
  \label{fig:cost-vs-latency}
\end{figure*}

We found two keys to optimize LLMs for inference efficiency.
First, we found it useful to build a powerful and abstract \textit{partitioning framework} to enable reaching the limits of model parallel scaling given the limited parallelizability of Transformer inference.
Within this framework, we analytically solve for the best partitioning strategy for a given model size with specific application requirements. 
This enables the user to intuitively understand the tradeoffs and select the best multi-axis tensor partitioning strategy, batch size and chip configuration for their application, in contrast to a black-box exhaustive search over partitioning strategies~\cite{zheng2022alpa, xu2021gspmd}. 
To fully realize the performance in practice, we use additional fine-grained control over cross-chip collective operations and low-level scheduling optimizations. 
Second, we apply memory optimizations and take full advantage of PaLM's multiquery attention to reduce unnecessary tensor overheads and maximize the batch size that fits on a given number of chips, enabling higher throughput.

The primary goal of this paper is to provide a set of engineering principles for how best to partition a model in order to scale Transformer inference. In other words, how is the performance of different partitioning strategies affected by changes in model size, sequence length, and number of hardware chips? How does the optimal partitioning strategy change when trading off between latency and throughput? What is the intuitive and mathematical reasoning behind these effects? As we show in later sections, the right tradeoffs and strategies change as model size, sequence length, and application requirements for latency and throughput targets change, so having a framework that enables easy expression of different strategies and choices is important.

In Section 2, we describe the specific metrics and tradeoffs we use to compare different partitioning strategies. In Section 3.1, we provide an overview of partitioning principles for large language models. In the remainder of Section 3, we describe a number of specific partitioning strategies, with an empirical validation on the PaLM family of large language models in Section 4. %

For a state-of-the-art 540B parameter dense model running on 64 TPU v4 chips, we achieve a low-batch-size latency of 29ms per token during generation (with int8 weight quantization) and a 76\% MFU during large-batch-size processing of input tokens while supporting a large context length of 2048 tokens. 
Figure~\ref{fig:cost-vs-latency}(left) shows our performance for generating text using the PaLM models. For an interactive application such as a chatbot running on PaLM 540B with int8 weights, our implementation on 64 TPU v4 chips can process 64 tokens of text from a user, consult a cached conversation history of 1920 tokens, and generate a 64-token response in a total of 1.9 seconds. 
For an offline throughput-oriented application, our implementation can process 1984 tokens of input and generate 64 tokens of output, for huge numbers of examples, with an overall FLOPS efficiency of 73\%. Table~\ref{table:cost-vs-latency} shows more details on a few specific scenarios.

\section{Inference Cost Tradeoffs}\label{sec:analyticalmodel}
Scaling up model sizes can unlock new capabilities and applications but has fundamental tradeoffs in terms of inference cost. We measure the inference cost in terms of the following metrics: latency, throughput, and model FLOPS utilization. The \emph{latency} is the total time for an inference and can be broken down into the time to process the input tokens present at the start of the inference (which we call ``prefill'') and the time to autoregressively generate output tokens (which we term ``decode''). The decode latency can also be measured ``per step'', i.e. divided by the number of tokens in each sequence. The \emph{throughput} of prefill or decode is the number of tokens processed or generated per second. The \emph{model FLOPS utilization (MFU)} is the ratio of the observed throughput to the theoretical maximum throughput if the benchmarked hardware setup were operating at peak FLOPS with no memory or communication overhead.

Larger models do not fit on a single accelerator chip and need to be partitioned across many accelerator chips to fit in memory. This also enables us to divide the memory and compute costs described below over all the chips, but comes at the cost of introducing chip-to-chip communication.

\paragraph{Memory costs.}
We store tensors such as weights and the KV cache in on-device high-bandwidth memory (HBM).
While there are other tensors that pass through the HBM, their memory footprint is much smaller, so we focus on just these two largest groups of tensors.
These tensors need to be transferred from HBM to the compute cores of the chip once per forward pass (prefill or decode step) of the model.
This takes a certain amount of time, which we call the ``memory time.''
At small batch sizes and sequence lengths, the time to load weights dominates.
At larger batch sizes and sequence lengths (e.g.\ 2048+ tokens with batch size 512+), the time to load the KV cache dominates.

\paragraph{Compute costs.} 
An $N$-parameter decoder-only model requires $2N$ matmul FLOPs in the forward pass per token seen because each matmul performs one multiplication and one addition per pair of input token and parameter values in the forward pass~\citep{kaplan2020scaling}. If all chips were running at peak FLOPS, these matmuls would take a certain amount of time, which we call the ``compute time.''
The matmuls in the attention mechanism typically add a much smaller number of FLOPs per token for large models and can often be excluded.
Even though the computational cost of attention is relatively small, it can still account for a significant fraction of memory capacity and bandwidth costs, since (unlike the weights) the KV cache is unique for each sequence in the batch.

\subsection{Expected tradeoffs and challenges}
Both the weight loading part of the memory time and the non-attention compute time are proportional to the model size and inversely proportional to the number of chips.
However, for a given partitioning layout, the time needed for chip-to-chip communication decreases less quickly (or not at all) with the number of chips used, so it becomes an increasingly important bottleneck as the chip count grows. 
We consider some scenarios where these tradeoffs become especially challenging.

If an application requires the \emph{lowest possible latency}, we need to apply more chips and partition the model in as many ways as we profitably can. Lower latency can often be achieved with smaller batch sizes, but smaller batch sizes also result in worse MFU, resulting in a higher total cost (in terms of chip-seconds or dollars) per token. %

If an application requires generating text with \emph{long attention contexts}, it substantially increases the inference time. For a 500B+ model with multihead attention, the attention KV cache grows large: for batch size 512 and context length 2048, the KV cache totals 3TB, which is 3 times the size of the model's parameters. The on-chip memory needs to load this KV cache from off-chip memory once for every token generated during which the computational core of the chip is essentially idle. %

If an applications requires \emph{offline inference} and latency is not a concern, the primary goal is to maximize per-chip throughput (i.e., minimize total cost per token). It is most efficient to increase the batch size because larger batches typically result in better MFU, but certain partitioning strategies that are not efficient for small batch sizes become efficient as the batch size grows larger. %

\subsection{Inference Setup}
We briefly introduce the inference setup and notation. We consider a Transformer model with $\nparams$ parameters laid out for inference on $\nchips$ chips. The model has
model (or embed) dimension $\dmodel$ (or $E$), feedforward intermediate dimension $\dff$ (or $F$), and $\nheads$ (or $H$) heads.

Each example in a batch of $B$ sequences has $\Linput$ tokens of input text, and generates $\Lgen$ tokens of output text. Since the input tokens are all present at the start of the inference, we can run the
model over all $B\times \Linput$ many tokens in parallel, in a single forwards pass
over all the tokens. We call this step \emph{prefill}. The output tokens are generated autoregressively, with a sequential loop of
$\Lgen$ steps. Each step consists of a single forwards pass through the model, after
which we sample one new token for each of the $B$ examples in the batch. This
loop is known as \emph{generation} or \emph{decode}.

Since prefill can run in parallel over $\Linput$, but decode must run sequentially over $\Lgen$, the two phases have different performance characteristics and we analyze them separately.

\section{Partitioning for inference efficiency}\label{sec:design}
We must partition large models over many chips in order to fit weight and activation tensors in memory and fit compute and memory time within latency requirements.
Model partitioning introduces communication between chips, and different partitioning strategies for a given model involve different patterns and amounts of communication. In this section, we detail several high-level strategies for partitioning a large Transformer language model for cost-effective and latency-effective inference.

\subsection{Partitioning notation and communication collectives}
We describe the partitioning layouts in this 
section based on a TPU v4 system with 3D torus topology $X\times Y\times Z$. 
Following \cite{xu2021gspmd}, we use subscripts to specify the 
tensor dimension that is partitioned. For example, notation $BLE_{xyz}$
means that the last dimension $E$ of a tensor of logical shape $BLE$ is split into $X\times Y\times Z$ partitions, where $x$,
$y$ and $z$ refer to the physical TPU v4 axes, and 
the per-chip tensor is of shape $[B, L, E/(X\times Y\times Z)]$. 
Here $B$, $E$ and $F$ refers to batch, model embed and MLP feedforward dimension. We use $L$ to refer to the sequence length and explicitly specify prefill or generation phase. 

If a tensor is replicated over an axis $x$, that axis is omitted from the notation. We also use a suffix ``partialsum-$x$'' to indicate that a given tensor has been contracted (summed) locally on each chip (over some axis not represented in the shape), but still needs to be summed across the chips in the TPU $x$ axis (creating a tensor replicated over $x$) before the result is meaningful.

We use several communication collectives originating from MPI~\cite{clarke1994mpi}. The \emph{all-reduce($x$)} primitive sums a partialsum tensor such as $BLE_{yz}( \textrm{partialsum-}x)$ across sets of chips in the $x$ axis of the torus and broadcasts the sum back to all the involved chips, returning output of shape $BLE_{yz}$. For the reasons outlined in \citet{rajbhandari2020zero}, we typically split all-reduce into two phases: a reduction phase and a broadcast phase. The reduction phase is called \emph{reduce-scatter($x$)}, and it sums $BLE_{yz}(\textrm{partialsum-}x)$ tensors across sets of chips in the $x$ axis but produces an output that's sharded rather than replicated over the chips in that axis, in a layout such as $B_xLE_{yz}$ or $BLE_{xyz}$. The  broadcast phase is called \emph{all-gather($x$)}, and it broadcasts and concatenates the tensor $BLE_{xyz}$ to all chips in the $x$ axis, producing an output $X$ times larger than its input, replicated over the $x$ axis: $BTE_{yz}$. 
The \emph{all-to-all} collective shifts sharding from one tensor dimension to another, e.g. $BLH_x Q \to B_x LHQ$ by using direct communication between every (source, destination) pair. Figure~\ref{sec:appendix-cost-allreduce} illustrates these primitives.

\subsection{Partitioning the feedforward layer}
\subsubsection{Feedforward layer, 1D weight-stationary layout}
\paragraph{Overview.} When a model doesn't fit on a single chip, the simplest partitioning strategy is \textit{1D weight-stationary}, where each $E \times F$ weight matrix is partitioned (or sharded) among $\nchips$ along the $E$ \textit{or} $F$ axis. Each weight shard is multiplied by the appropriate activation shard on each chip, and the results are aggregated between the chips with an all-gather and/or reduce-scatter. Additionally, when computing two consecutive matrix multiplications (as in a Transformer MLP block), there is a ``trick'' \cite{shoeybi2019megatron} to avoid any cross-chip communication between the matmuls: if the first matmul is partitioned by the output axis, the resulting activation shard on each chip will be the exact one needed to compute the second matmul partitioned by the input axis.

As we parallelize the computation across more chips, the memory latency and compute latency does decrease, often near-linearly. However, the communication latency remains roughly constant independent of the number of chips used, since the entire activation matrix is aggregated across chips for every pair of matrix multiplications. As the number of chips grows larger, communication becomes a bottleneck.

\paragraph{Details.} We consider as a baseline the layout 
where the weights and activations of the feedforward layer are 
partitioned over $\nchips$ along the $\dff$ dimension, 
as in Megatron~\cite{shoeybi2019megatron}. Figure~\ref{fig:ffn}(a) shows the partitioning layout for this case. 
On the TPU v4's 3D torus topology the partition layout for weights is $EF_{xyz}$ and $F_{xyz}E$, i.e. 
they are partitioned in to $X\times Y \times Z = \nchips$ partitions 
with $X$, $Y$, and $Z$ partitions across physical TPU axes. The weights are kept stationary 
in each chip, and the activations are transferred between chips to match the weight layout,
requiring one all-gather and one reduce-scatter.

In this 1D weight-stationary partitioning strategy, 
each chip gets inputs and outputs of shape $BLE$ in the reduce-scatter and all-gather respectively. We derive the the communication cost of these operations in Appendix~\ref{sec:appendix-cost-allreduce}. The resulting communication time is
\begin{align*}
    T_\textrm{comm} &= \dfrac{2BLE}{ \textrm{network bandwidth} }.
\end{align*}

\subsubsection{Feedforward layer, 2D weight-stationary layout}
\label{sec:weight-stationary}

\paragraph{Overview.} For a larger number of chips, a more economical strategy involves partitioning each $E \times F$ weight matrix along both the $E$ \textit{and} $F$ axes, such that each shard is roughly square. For example, if $E = 1024$, $F = 4096$, and $\nchips = 64$, then we would shard 4-ways among $E$ and 16-ways among $F$, so that each of the 64 chips stores a 256-by-256 chunk of the weight matrix, and activations are transferred between chips. This is called \textit{2D weight-stationary}. The total compute cost is the same as 1D weight-stationary, but communication is much more efficient: when multiplying an activation matrix through a set of consecutive weight matrices, we can \textit{alternate} which of the two axes we perform the activation aggregation on between each multiplication. With the correct partitioning, each chip will always have the necessary activation shard to multiply with its weight shard, without ever having a fully replicated copy of the activation tensor. Since each axis is partitioned on $O(\sqrt{\nchips})$, the communication time scales as $O(\frac{1}{\sqrt{\nchips}})$ rather than remaining constant.
This means that even if the 2D layout is communication-limited at a certain chip count and batch size, we can continue to reduce latency by adding more chips, because communication time continues to reduce.

However, while the 1D weight-stationary ``trick'' requires us to only aggregate over the $\dmodel$ dimension, 2D weight-stationary requires alternating aggregation over the $\dmodel$ and $\dff$ dimensions. Therefore, 2D weight-stationary becomes more communication-efficient when $\sqrt{\nchips} > \frac{\dff}{\dmodel}$. Since typically $\dff = 4\dmodel$, this occurs when $\nchips > 16$. %

\begin{figure}[tb!]
\centering
\includegraphics[height=2.6in]{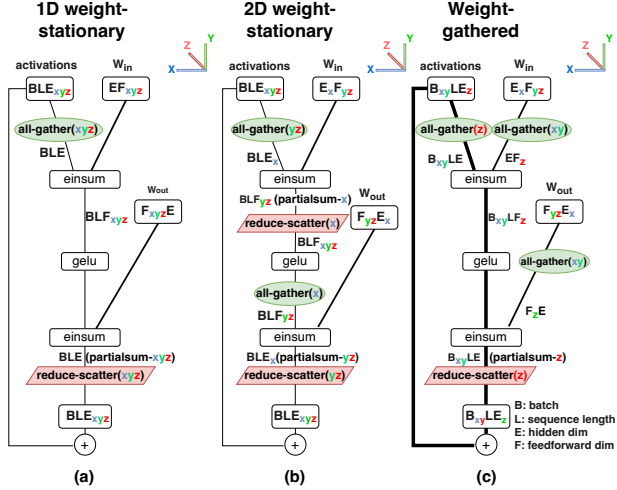}
\caption{Partitioning layouts for feedforward layer.}
\label{fig:ffn}
\end{figure}

\paragraph{Details.} Figure~\ref{fig:ffn}(b) shows the partitioning layout. Whereas the 1D weight-stationary layout runs its all-gather and reduce-scatter with unsharded shape $BLE$ per chip, this 2D weight-stationary layout partitions $\dmodel$ so that the communication volume for $\dff$ partitioning is reduced from $BLE$ to $\frac{BLE}{X}$. This comes at the cost of introducing a second pair of reduce-scatter and all-gather operations, whose cost must be balanced with the existing communication.

The partitioning layout for weights is $E_{x}F_{yz}$, 
i.e. they are partitioned along the $\dmodel$ dimension into $X$ partitions and 
along the $\dff$ dimension into $Y\times Z$ partitions, 
where $X \times Y\times Z = \nchips$. The partitioning layout for the input activations
is the same as the previous section. 
Note that we again keep 
the partitioned weights stationary on their chips, but because of their 2D layout,
the activation communication includes two all-gathers and reduce-scatters.

We derive the optimal values of $X$, $Y$ and $Z$ to minimize total communication time in Appendix~\ref{sec:appendix-partitioning-ws2d}. Assuming $\dff=4 \times \dmodel$, 
we achieve the minimum communication time with $X=0.5\times\sqrt{\nchips}$ 
and $YZ=2\times\sqrt{\nchips}$. The resulting total communication time is:
\begin{align*}
    T_\textrm{comm} &= \dfrac{8 B L E}{\sqrt{\nchips}\times \textrm{network bandwidth}}.
\end{align*}

\subsubsection{Feedforward layer, weight-gathered layout}
\label{sec:weight-gathered}
\paragraph{Overview.} In the previously described weight-stationary strategies, each chip stores one shard of each weight matrix in memory, and that chip is responsible for multiplying it's ``stationary'' weight shard with each corresponding activation shard. The output of each per-chip matrix multiplication must then be aggregated between chips to be used as input to the subsequent operations.

However, as the batch size (and sequence length) grows larger, the size of the output activations may become significantly larger than the size of the weights. When this happens, it can become more economical to keep the activations stationary on each chip, and instead transfer the weights between chips. For very large batch sizes, it is best to keep the activations fully stationary between sequential matrix multiplications, requiring that we fully transfer the weights between all chips. We call this approach \textit{XYZ-weight-gathered}. For moderate batch sizes, it is beneficial to use a ``hybrid'' approach where both weights and activations are partially transferred along different axes. We refer to these approaches as \textit{X-weight-gathered} and \textit{XY-weight-gathered}.

\paragraph{Details.} Figure~\ref{fig:ffn}(c) shows the XY-weight-gathered layout. A key aspect of the specific layout we choose is that weights start in the same $E_xF_{yz}$ layout as in 2D weight-stationary, so that we can use the same weight layout for weight-gathered (duing prefill) and weight-stationary (during decoding). Just before the einsums, the weight tensors are all-gathered over the $X$ and $Y$ axes, with communication volume $\frac{EF}{Z}$. This is additional communication relative to weight-stationary layout, but in return we reduce the communication on activations: one reduce-scatter/all-gather pair for activations is skipped, and the communication volume on the other pair drops from $\frac{BLE}{X}$ to $\frac{BLE}{XY}$.

By changing the relative sizes of the $X$, $Y$, and $Z$ axes, we can trade off weight communication against activation communication, and thereby minimize the total communication volume. But we choose to share the weights between weight-stationary and weight-gathered layouts, which means we are required to match the choices of $X$, $Y$ and $Z$ made for the weight-stationary layout. What we do instead is pick between a few variants of the weight-gathered layout. The variant shown in Figure~\ref{fig:ffn}(c) uses all-gather($xy$) for the weights and $B_{xy}LE_z$ partitioning of batch for the activations. Our other variants use all-gather($x$) or all-gather($xyz$) for weights, and correspondingly use $B_xLE_{yz}$ or $B_{xyz}LE$ partitioning of the activations. Figure~\ref{fig:ffnweightgathered} shows the three weight-gathered layouts.

\begin{figure}[t]
\centering
\includegraphics[width=3.0in]{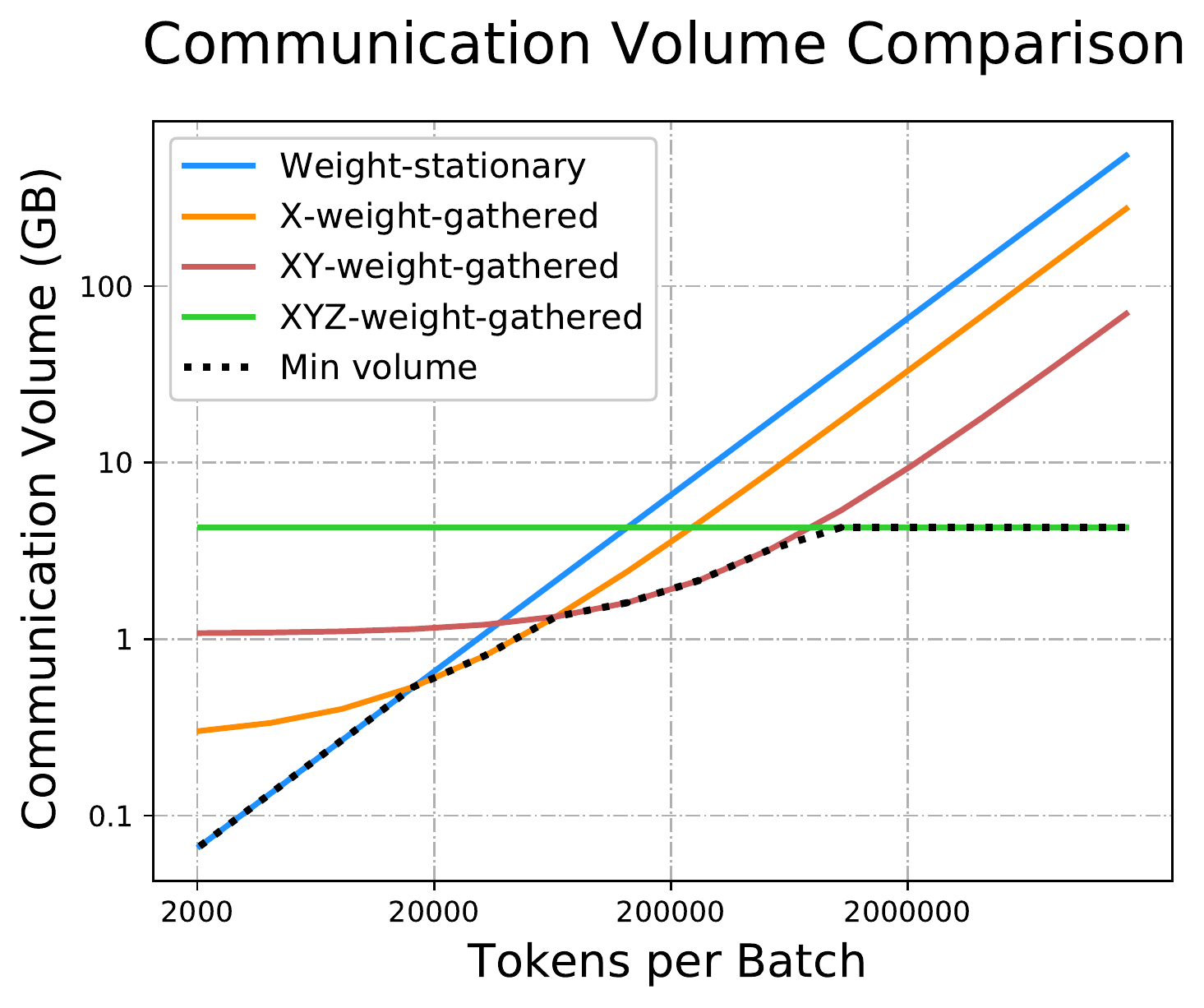}
\vspace{-0.2in}
\caption{Communication volume as a function of batch size, for a feedforward layer. As batch size (in tokens) grows, it is better to switch to a layout that all-gathers the weights over increasingly more chips to minimize the communication volume. Communication volumes estimated for $X=Y=Z=4$, $\dmodel=16384$, and $\dff=65536$. }
\label{fig:communication-volume-by-batch}
\vspace{-0.2in}
\end{figure}

Figure~\ref{fig:communication-volume-by-batch} shows how the communication-optimal configuration switches between these layouts as batch size grows -- while the 2D weight-stationary strategy minimizes communication at low tokens per batch, different weight-gathered layouts are optimal at larger number of tokens per batch. This highlights the importance of choosing different inference configurations depending on application goals.

We now show the asymptotic scaling of weight-gathered layouts. Let $N$ be the number of chips that weights are all-gathered over: $N=X$ in $X$-weight-gathered, $N=XY$ in $XY$-weight-gathered, $N=XYZ$ in $XYZ$-weight-gathered. Total communication is minimized by the choice $N=\sqrt{\frac{BL\nchips}{F}}$ which we derive in Appendix~\ref{sec:appendix-weight-gathered}. The total communication time is
\begin{align*}
T_\textrm{comm} & = 4 E \dfrac{\sqrt{B L F}}{\sqrt{\nchips}\times \textrm{network bandwidth}}
\end{align*}
Note that $B L$ corresponds to the total batch size in tokens. 
The communication time for the weight-stationary layout is linear in $B L$, while 
the communication time for the weight-gathered layout is linear in $\sqrt{B L}$. 
Therefore, the weight-gathered layout becomes cheaper when the batch size and prefill sequence length are sufficiently large.

\begin{figure}[tb!]
\centering
\includegraphics[width=0.95\columnwidth]{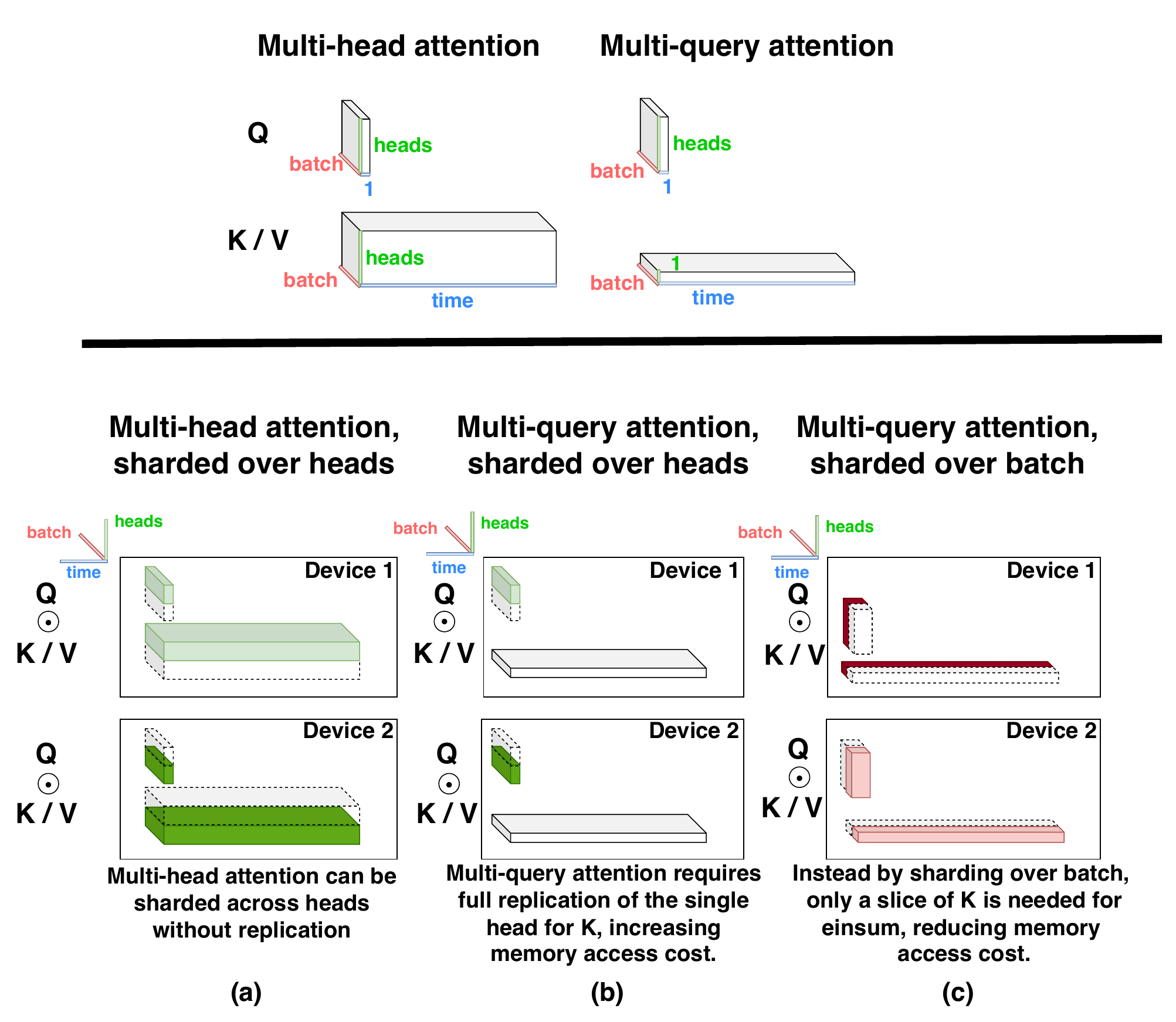}
\caption{Multiquery attention has lower memory cost to load the KV cache when sharded over batch.}
\label{fig:mqa}
\end{figure}

\subsection{Partitioning the attention layer}
\label{sec:multiquery}
Multihead attention can be parallelized in essentially the same ways as a feedforward layer, with $\nheads$ replacing $\dff$.
But inference with multihead attention incurs significant memory capacity and bandwidth costs to store and load the KV cache, and these costs can dominate the rest of the inference at large batches or long context lengths.

An alternative approach, called \textit{multiquery attention}~\cite{shazeer2019fast,chowdhery2022palm}, still emits $\nheads$ for the query tensor, but only a single head for the key and value tensors. This key and value head is shared across the $\nheads$ query heads.
This reduces the size of the KV cache tensors by a factor of $\nheads$ and hence the memory time spent loading them. But it also removes an axis otherwise used for parallelism, so the KV cache and related computations need to be partitioned differently.

\paragraph{Partitioning strategy.} The key design consideration is to minimize the memory time of repeatedly loading the KV cache that dominates the inference cost.
The partitioning layout of projection matrices that have a $\nheads$ dimension ($W_Q$ and $W_O$ in multiquery attention, and those two plus $W_K$ and $W_V$ in multihead attention) should match the layout used in the feedforward layer.

Figure~\ref{fig:mqa}(a) shows a typical partitioning layout for multihead attention, matching the 2D weight stationary feedforward layout. 
Here the $Q$, $K$, and $V$ activations are partitioned over the $\nheads$ dimension into $\nchips$ partitions when $\nheads$ is a multiple of $\nchips$. 
For $\nchips$ greater than $\nheads$, the attention heads are partially replicated. 
The most similar partitioning layout for multiquery attention (shown in Figure~\ref{fig:mqa}(b)) treats the KV cache the same as in multihead attention.
Even though the key and value tensors are shared across all heads, they must be replicated on each chip and the memory cost savings of multiquery attention are lost.

\begin{figure}[tb!]
\centering
\includegraphics[height=2.6in]{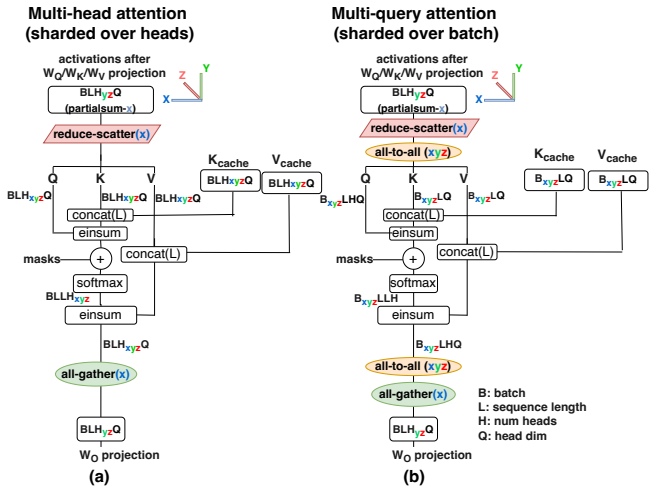}
\caption{Comparison of partitioning layouts for attention layer: multihead attention  sharded over heads versus multiquery attention sharded over batch.}
\label{fig:attn}
\vspace{-0.2in}
\end{figure}

We instead propose a partitioning strategy for the multiquery attention where the $Q$, $K$, and $V$ matrices are partitioned over the batch $B$ dimension into $\nchips$ partitions. 
Figure~\ref{fig:mqa}(c) shows that this reduces the memory cost of loading the KV cache per chip by a factor of $\nchips$, thereby reducing the memory time by the same factor. 
The proposed partitioning strategy incurs additional communication cost of resharding the input activation tensors using an all-to-all collective as shown in Figure~\ref{fig:attn}(b) in comparison to the multiquery attention sharding strategy shown in Figure~\ref{fig:attn}(a) where the $Q$, $K$, and $V$ matrices are partitioned over the heads dimension. %

During autoregressive generation, there is only one token per example of Q, K, and V tensors, whereas the KV cache has many (perhaps 2048) tokens. Since the KV cache is orders of magnitude larger than the Q, K, and V tensors, it is very profitable to spend the all-to-all communication time on the small tensors to save the memory time on the large tensors.

During prefill, it is typically not profitable to shard attention over batch. The Q tensor has many (perhaps 2048) tokens, all of which are queried against the same K and V tensors. The memory load of the K and V tensors is amortized over all tokens in the Q tensor, and so this memory load is typically not a bottleneck during prefill. Therefore for prefill we use the sharded-over-heads layout.

With the proposed partitioning layout,
multiquery attention enables using larger
batch sizes and sequence lengths, thereby increasing throughput in addition to the latency reduction from reduced memory time.
As shown in Section~\ref{sec:attention-ablation}, the savings are an order of magnitude compared 
to multihead attention.

\subsection{Parallel attention/feedforward layers}\label{sec:parallel-layers}
We discuss the inference latency gains from the ``parallel'' formulation of each Transformer block~\citep{gpt-j} as used in PaLM~\cite{chowdhery2022palm} instead of the standard ``serialized'' formulation, where the feedforward layer and attention layer are computed in parallel from the layernormed input and summed to get the output.

The benefits from the parallel formulation are as follows. 
First, there is only one layernorm per layer instead of two, which reduces latency at small batch sizes. 
Second, the input matrices of the feedforward layer can be fused with the query projection matrix $W_Q$  
of the attention layer, the key/value projection matrices $W_K$ and $W_V$ can be fused in the attention layer, 
and the output matrix of the feedforward layer can be fused with the output projection matrix 
$W_O$ of the attention layer. This fusion results in higher FLOPS utilization because 
larger matrix-multiplications run more efficiently on accelerators. 
More importantly, it also eliminates one of the two all-reduce operations in each Transformer layer needed for $\dff$/$\nheads$ parallelism, cutting communication time over this axis in half.

\subsection{Low-level optimizations}\label{sec:lowlevel}

We use the Looped CollectiveEinsum technique from \cite{wang2023collectivematmul} to run communication concurrently with computation. 
This allows us to partially or fully hide the communication time of most of the reduce-scatter and all-gather operations in Figures~\ref{fig:ffn} and~\ref{fig:attn}. For all reduce-scatter operations in Figures~\ref{fig:ffn} and~\ref{fig:attn}, we had a choice of whether to reduce-scatter into a batch or sequence dimension ($B$ or $L$) or into the hidden dimension ($E$ or $F$). We chose the latter, because it exposes more effective opportunities for Looped CollectiveEinsum, whereas \citet{korthikanti2022reducing} chose the former, to avoid communication in layernorm.

The CollectiveEinsum loops are the overwhelming majority of the inference latency, so we invested considerable effort to maximize their performance. First, we used the underlying ``async CollectivePermute'' APIs of \citet{wang2023collectivematmul} to develop a suite of variants of the CollectiveEinsum concept, to optimize for different scenarios: latency versus throughput, different numbers of torus axes, fusing with different input/output collectives. Second, we explicitly match up communication collectives with the matrix multiplies that they should be fused with, to maximize the potential for overlap. Through such optimizations, we achieved about 1.4 times better performance than the simpler compiler-partitioned-and-scheduled implementation that we started with. Some of the weight-gathered layouts would exhaust memory without these optimizations.%

We also included the following low-level optimizations: better in-memory layout of tensors to minimize padding and copying during matrix multiplies, faster top-$k$/top-$p$ implementations for decode sampling, faster log-base-2 implementations of Softmax and Swish, and support for incremental processing of sequences during prefill~\cite{fastertransformer}.
\subsection{Quantization}
\label{sec:incrementality}
We use the AQT library~\cite{aqt2022github} to reduce the memory cost of 16-bit weights by converting them to int8 without noticeable quality loss. This enables memory time savings from weight loading, which is especially helpful in the low batch size regime, and it reduces communication volume in weight-gathered layouts. We have not implemented \emph{activation} quantization~\cite{abdolrashidi2021pareto}, but we are hopeful that it could reduce compute time in large-batch configurations and reduce communication volume of activations in weight-stationary layouts.

\section{Case Study for PaLM Models}\label{sec:evals}
\paragraph{Methodology} We now conduct an empirical study of our techniques on the PaLM family of models \cite{chowdhery2022palm}, which we select since the model architecture incorporates the techniques of multiquery attention and parallel attention and feedforward layers.

\begin{figure}[t]
\centering
\includegraphics[height=2in]{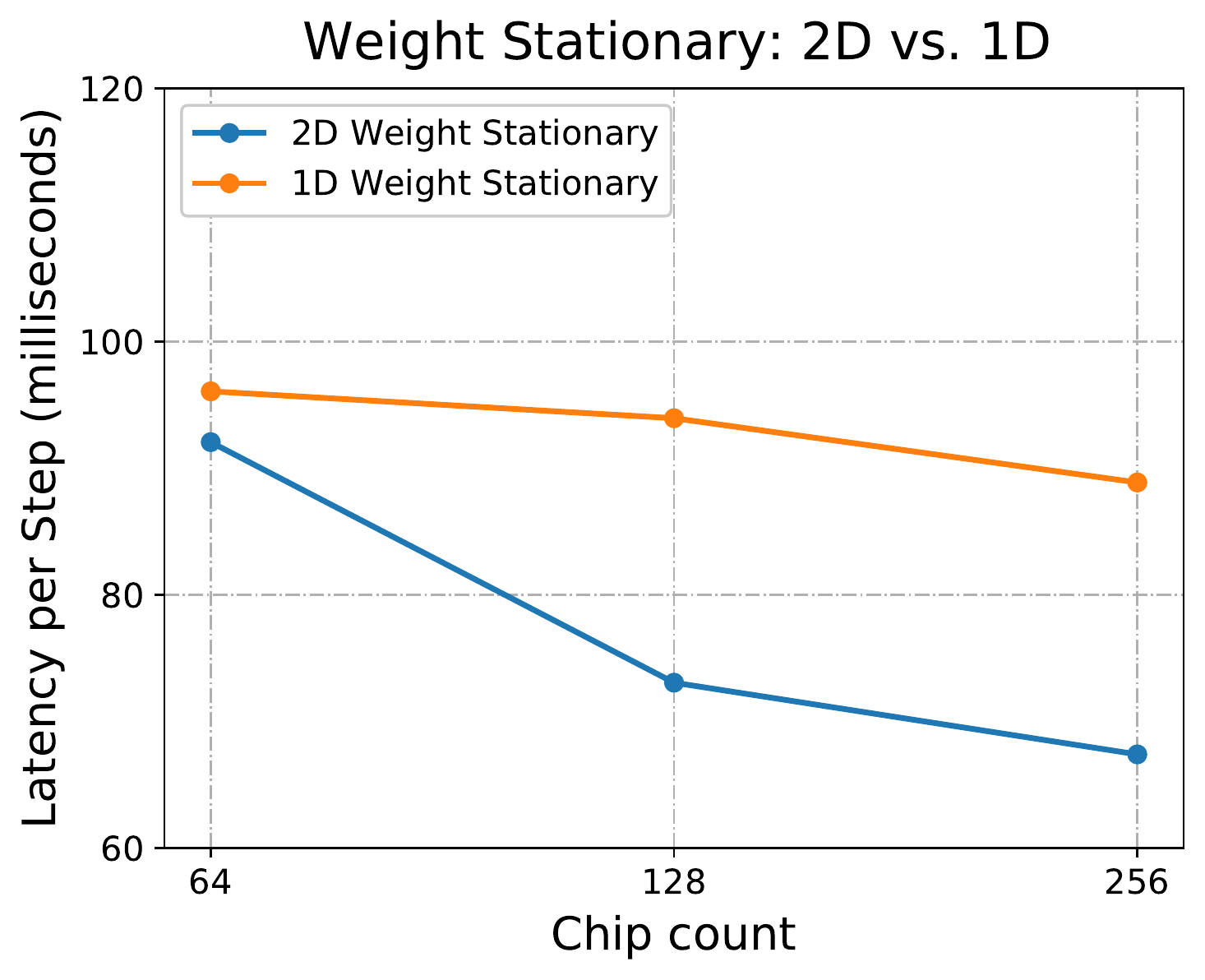}
\caption{Latency per token doing text generation of PaLM 540B for 2D and 1D weight stationary layouts on 64 chips.}
\label{fig:ffn-1d-vs-2d}
\vspace{-0.2in}
\end{figure}

Our inference framework is based on JAX~\cite{jax2018github} and XLA~\cite{XLA}, and our original high-level implementation was based on T5X~\cite{t5x2021github}. We use up to 256 TPU v4 chips~\cite{googlecloudtpu} for our benchmarks. Each TPU v4 chip can run bfloat16 matrix arithmetic at 275 TFLOPS, has 32 GiB of High Bandwidth Memory (HBM) at 1200 GB/s of bandwidth, and has 270 GB/s of interconnect bandwidth in a 3D torus topology~\cite{TPUv4}.

For the PaLM 540B model we padded the number of attention heads up from 48 to 64 in order to partition more effectively on 64+ chips. This adds 18B parameters to the model, which comes at a 3\% MFU cost, which was more than recovered by being able to partition more effectively.

\subsection{Partitioning feedforward layer}\label{sec:eval_ffn}
We evaluate the relative performance of our feedforward layer partitioning strategies. First we evaluate performance of decoding. We use batch size 512 to balance latency and MFU. Figure~\ref{fig:ffn-1d-vs-2d} shows the performance of 1D and 2D weight-stationary layouts as we increase the chip count. Both layouts start to become communication-limited, but the 2D layout performs better because of its asymptotically better scaling with chip count.

\begin{figure}[t]
\centering
\includegraphics[height=2in]{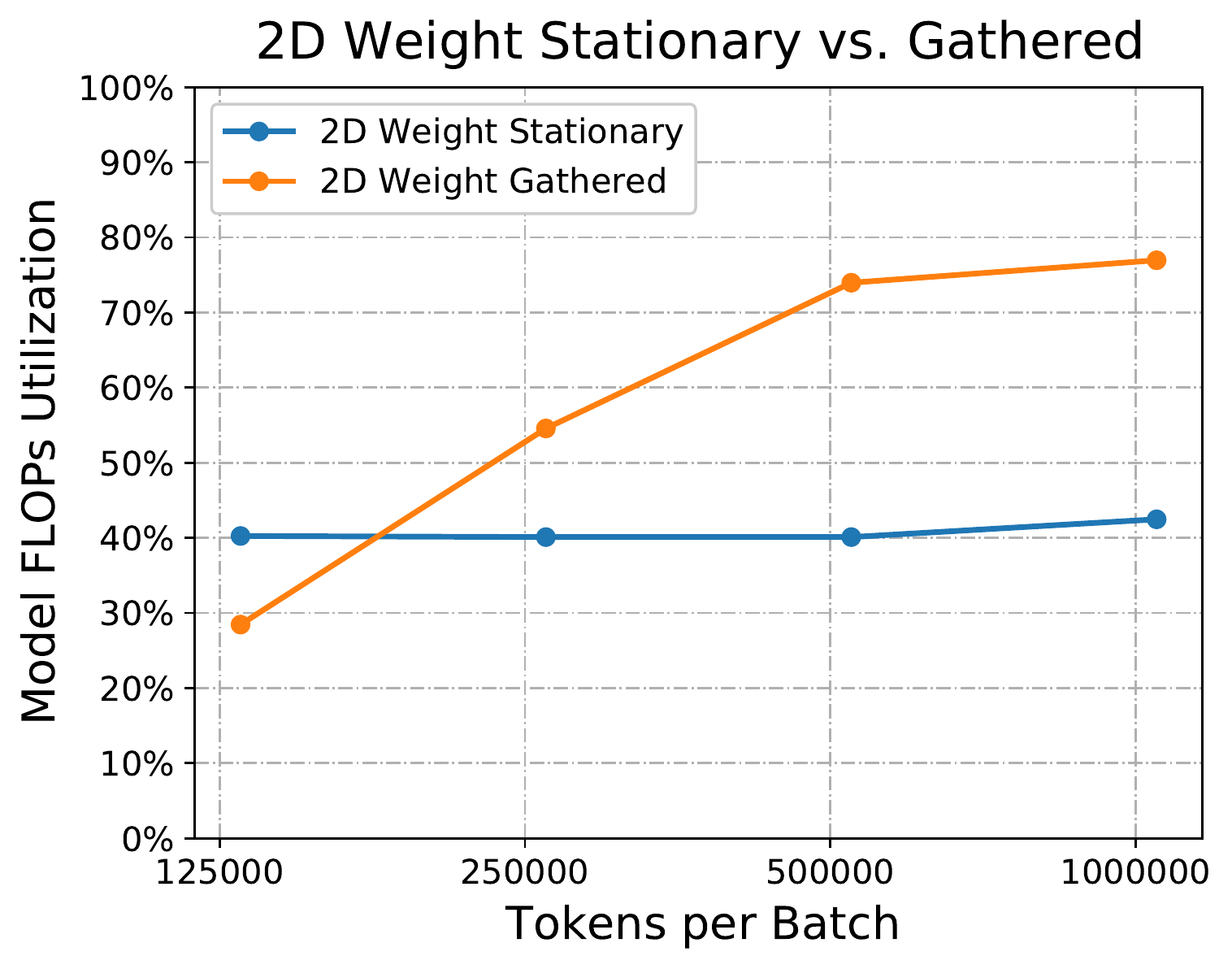}
\caption{Model FLOPS utilization running prefill on PaLM 540B on 64 chips, with sequence length 2048. We report batch size measured in tokens: number of sequences multiplied by sequence length. As batch size (in tokens) grows, note that it is better to switch from the 2D weight stationary to the weight gathered approach to improve MFU.}
\label{fig:ffn-prefill-batch-size-sweep}
\vspace{-0.1in}
\end{figure}

Next we consider the prefill phase. We consider batch sizes from 2048 tokens (1 example, 2048 tokens) to 1 million tokens (512 examples, 2048 tokens per example). Figure~\ref{fig:ffn-prefill-batch-size-sweep} shows that the optimal partitioning layout switches from the 2D weight-stationary layouts to the weight-gathered layouts as the batch size increases. The weight-gathered layouts are inefficient at low batch sizes, but eventually they become the most efficient at high batch sizes, achieving 76\% MFU when the communication overhead is almost negligible. Such large batch sizes would fail from memory exhaustion without multiquery attention, as shown in Section~\ref{sec:attention-ablation}. This highlights the importance of flexibility in configuring the inference system with different choices depending on the application setting and goals.

These results give us our basic strategy for selecting partitioning layout: during the prefill phase, we select from weight-stationary and weight-gathered layouts based on the current number of tokens in the batch. During the generate phase, we select the 2D weight-stationary layout because the batch size in tokens is always small. 

\subsection{Partitioning Attention layer}\label{sec:attention-ablation}
We now evaluate the partitioning layout for multiquery attention proposed in Section~\ref{sec:multiquery}. We consider PaLM with multiquery attention in both the baseline layout that partitions by attention heads and the optimized layout that partitions by batch. We also create a modified variant of PaLM 540B which uses multihead attention instead of multiquery. To keep parameter count in the attention layer constant, we shrink $\dhead$ from 256 in the multiquery variant to 128 in the multihead variant.

At large batch sizes and context lengths, the KV cache can become very large, putting us at the risk of running out of memory. Table~\ref{table:attention-oom} shows that the optimized multiquery layout can fit up to 32--64 times longer context lengths than the multihead and baseline multiquery variant.

\begin{figure}[t]
\centering
\includegraphics[height=2in]{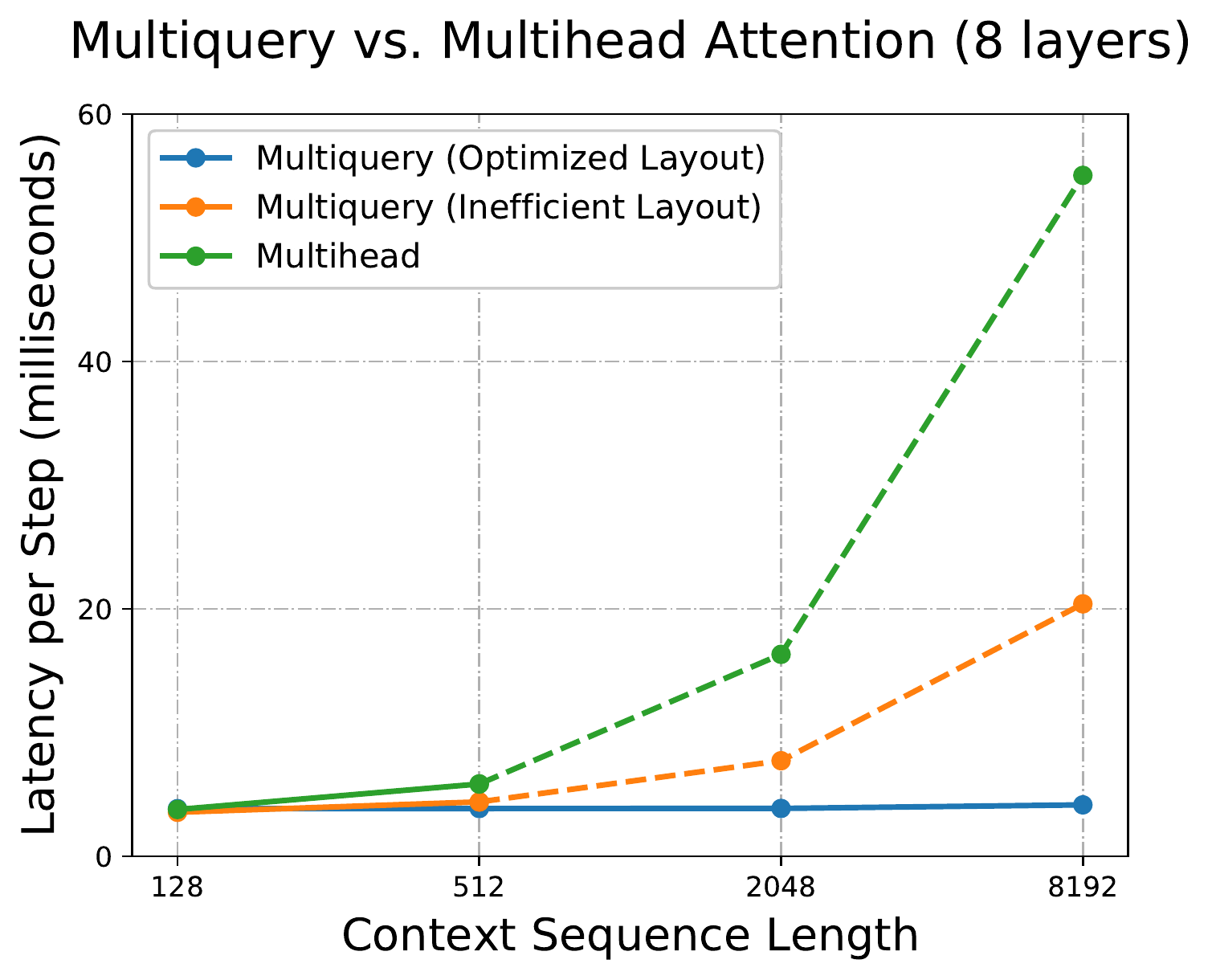}
\caption{Latency per generated token vs.\ sequence length, for an 8-layer version of PaLM 540B on 64 chips with batch size 256. The dotted line represents that on the full 118-layer model and context lengths longer than 512, the KV cache will not fit in memory when using multihead attention or the baseline multiquery partitioning.}
\label{fig:multiquery-ablation}
\vspace{-0.1in}
\end{figure}

During prefill, multiquery and multihead attention incur similar inference latencies because we compute many attention queries in parallel and the attention computation becomes compute-limited on the attention matrix multiplies. During generation, Figure~\ref{fig:multiquery-ablation} shows that the optimized multiquery layout improves speed. The speed improvement is small when the context length is short because almost all of the time is spent on the feedforward layer. As the context length grows longer, the time to load the KV cache in the attention layer becomes a much larger portion of overall inference time. Multiquery attention scales up to sequence lengths of 8192--32,768 tokens (batch sizes 512 and 128 respectively) with attention taking only 8--31\% of total runtime.

\begin{table}[t]
\centering
\small
\begin{tabular}{ c c c c}
\hline
 Model variant & $\dhead$ & \multicolumn{2}{c}{Max context length} \\
 & & batch=128 & batch=512 \\ 
 \hline
 Multihead & 128 & 1320 & 330 \\  
 Baseline multiquery & 256 & 660 & 165 \\    
 Optimized multiquery & 256 & 43,000 & 10,700\\ 
 \hline
\end{tabular}
\caption{Maximum context length supported for different attention variants of PaLM 540B on 64 chips. We reserve 30\% of the total memory for KV cache. Optimized multiquery attention enables up to 32x larger context lengths.}
\label{table:attention-oom}
\vspace{-0.2in}
\end{table}

\subsection{Parallel attention/feedforward layers}\label{sec:parallel-ablation}

We consider a variant of PaLM 540B with the parallel formulation of Transformer block replaced by serial attention/feed-forward layers. During generation, we use 2D weight-stationary layout, 64 chips, and batch size 512. The serial formulation incurs 14\% higher inference latency per step than the parallel version because of the increased communication time for activations. In the prefill phase, this difference shrinks because the weight-gathered layouts incur less activation communication.

\subsection{End-to-end results on PaLM}\label{sec:end-to-end-cost-vs-latency}
We find the Pareto frontier between efficiency and latency as we scale the model size for the PaLM family of models: 8B, 62B and 540B, with weights in either bfloat16 or int8. We use a context length 2048 and sweep over the batch size and chip count. 

To meaningfully compare throughput across multiple model sizes with different chip count and batch sizes, we report the \emph{cost} of an inference in terms of \emph{chip-seconds per token} calculated as
\[
    \textrm{cost (chip-seconds per token)} = \frac{\nchips \times \textrm{time}}{BL}.\]
This is directly proportional to operational cost and inversely proportional to MFU. %

Figure~\ref{fig:cost-vs-latency}(left) shows the relationship between model size, latency, and cost in the generate phase, at the Pareto frontier of optimal batch size, chip count, and partitioning strategy. The lowest cost is achieved at batch sizes larger than about 512, where the cost is proportional to the number of parameters. As we decrease the batch size, we improve the latency but incur higher cost per token. The minimum latency for generation is 3 times lower than the batch-512 latency. 

We observe that int8 weight quantization achieves the minimum latency in Figure~\ref{fig:cost-vs-latency} (left): for example, we achieve 28.5ms/token with int8 weights at batch size 64 on PaLM 540B, while we achieve 36.9ms/token with bfloat16 weights. At low latency targets the cost is improved just over a factor of 2, because low-batch-size cost is dominated by weight loading time. At large batch size, cost is more neutral between int8 and bfloat16, because large-batch cost is dominated by the compute time and the matmuls still use bfloat16 arithmetic. We believe that quantization of \emph{activations} to int8 could enable a further cost improvement.

Figure~\ref{fig:cost-vs-latency} (right) shows the relationship between model size, latency, and cost in the prefill phase. The tradeoff between batch size and latency is less severe in the prefill phase than the generate phase and even batch size 1 runs with fairly low cost. Further, the cost of batch-512 prefill is 2 times lower than batch-512 generate because of the increased MFU of the weight-gathered layouts we use during prefill. More details on the relationship between model size and MFU are presented in Figure~\ref{fig:mfu-vs-latency} and Section~\ref{sec:appendix-mfu-vs-latency} in the Appendix.

\begin{table}[t]
\centering
\small
\begin{tabular}{p{0.5in} l l l l}
\hline
 & \multicolumn{2}{c}{Low-latency} & \multicolumn{2}{c}{High-throughput} \\
 & Prefill & Decode & Prefill & Decode \\
 \hline
 Chips & 64 & 64 & 64 & 64 \\
 Batch & 1 & 64 & 512 & 512 \\
 FFN & WS 2D & WS 2D & WG XYZ & WS 2D \\
 Attention sharding & Head & Batch & Batch & Batch \\
 Weights format & int8 & int8 & bfloat16 & bfloat16 \\
 MFU & 43\% & 14\% & 76\% & 33\% \\
 Latency & 0.29s & 1.82s & 85.2s & 6.0s \\
 \hline
\end{tabular}
\caption{\small Example configurations for PaLM 540B, in the same setting as Figure~\ref{fig:cost-vs-latency}. Prefill latency is for processing 2048 tokens; decode latency is for generating 64 tokens. Feedforward network (FFN) layouts are Weight Stationary 2D (WS 2D, Section~\ref{sec:weight-stationary}) and Weight Gathered XYZ (WG XYZ, Section~\ref{sec:weight-gathered}). Attention layouts are from Section~\ref{sec:multiquery}.}
\label{table:cost-vs-latency}
\end{table}

\begin{table}[t]
\centering
\small
\begin{tabular}{p{0.5in} l l l l}
\hline
 & \multicolumn{2}{c}{Low-latency} & \multicolumn{2}{c}{High-throughput} \\
 & Prefill & Decode & Prefill & Decode \\
 \hline
 Chips & 16 & 16 & 32 & 8 \\
 Batch & 1 & 32 & 512 & 512 \\
 FFN & WS 2D & WS 2D & WG XYZ & WS 2D \\
 Attention sharding & Head & Batch & Batch & Batch \\
 Weights format & int8 & int8 & bfloat16 & bfloat16 \\
 MFU & 36\% & 8\% & 73\% & 37\% \\
 Latency & 0.16s & 0.73s & 20.2s & 5.1s \\
 \hline
\end{tabular}
\caption{\small Example configurations for PaLM 62B, in the same setting as Figure~\ref{fig:cost-vs-latency}. Prefill latency is for processing 2048 tokens; decode latency is for generating 64 tokens. Feedforward network (FFN) layouts are Weight Stationary 2D (WS 2D, Section~\ref{sec:weight-stationary}) and Weight Gathered XYZ (WG XYZ, Section~\ref{sec:weight-gathered}). Attention layouts are from Section~\ref{sec:multiquery}.}
\label{table:cost-vs-latency-62b}
\end{table}

Tables~\ref{table:cost-vs-latency} and~\ref{table:cost-vs-latency-62b} show some key configurations from the Pareto frontiers of Figure~\ref{fig:cost-vs-latency}, on PaLM 540B and PaLM 62B. In the low-latency scenarios we combine batch-1 prefill with batch 32-to-64 decode: batch size 1 achieves best latency in the prefill phase, but for the generate phase we can increase the batch size up to 64 with negligible latency impact, and doing so is dramatically better for generate MFU. This mixture of batch sizes is possible in practice either by generating multiple samples from the same input text, or by pipelining a batch-1 prefill server into a batch-64 decoding server. 

In the high-throughput scenarios of Tables~\ref{table:cost-vs-latency} and~\ref{table:cost-vs-latency-62b}, we use larger batch sizes and we switch partitioned layouts between prefill and decode. We use bfloat16 weights for high-throughput scenario, because the weight-loading time is unimportant at large batch sizes, and because our software is missing some optimizations for large-batch int8 mode. 

Comparing 62B (Table~\ref{table:cost-vs-latency-62b}) vs.\ 540B models (Table~\ref{table:cost-vs-latency}), we find that we use more chips for the 540B model, but similar batch sizes and the same partitioned layouts. High-throughput MFUs are similar between the model sizes. The low-batch-size latencies grow \emph{sublinearly} with model size at the Pareto frontier: even though larger models load proportionally more weights from memory, we can partition them across more chips before becoming communication-limited. We estimate an approximately square-root relationship between model size and latency based on Figure~\ref{fig:cost-vs-latency} (left). %

\section{FasterTransformer Benchmarks}\label{sec:faster-transformer}
We now compare with the FasterTransformer benchmarks~\cite{fastertransformer} across a wide range of batch sizes and configurations of prefill and generate. There are multiple differences between our benchmark setup and the FasterTransformer benchmark. In particular, we use different types of chips and chip counts -- FasterTransformer uses 16--32 NVIDIA A100s with 80GiB HBM, while we use 64 Google TPU v4 chips with 32GiB HBM. Therefore, we report throughput numbers in terms of MFU, which normalizes for both chip count and chip FLOPS.

\begin{figure}[t]
\centering
\includegraphics[height=2in]{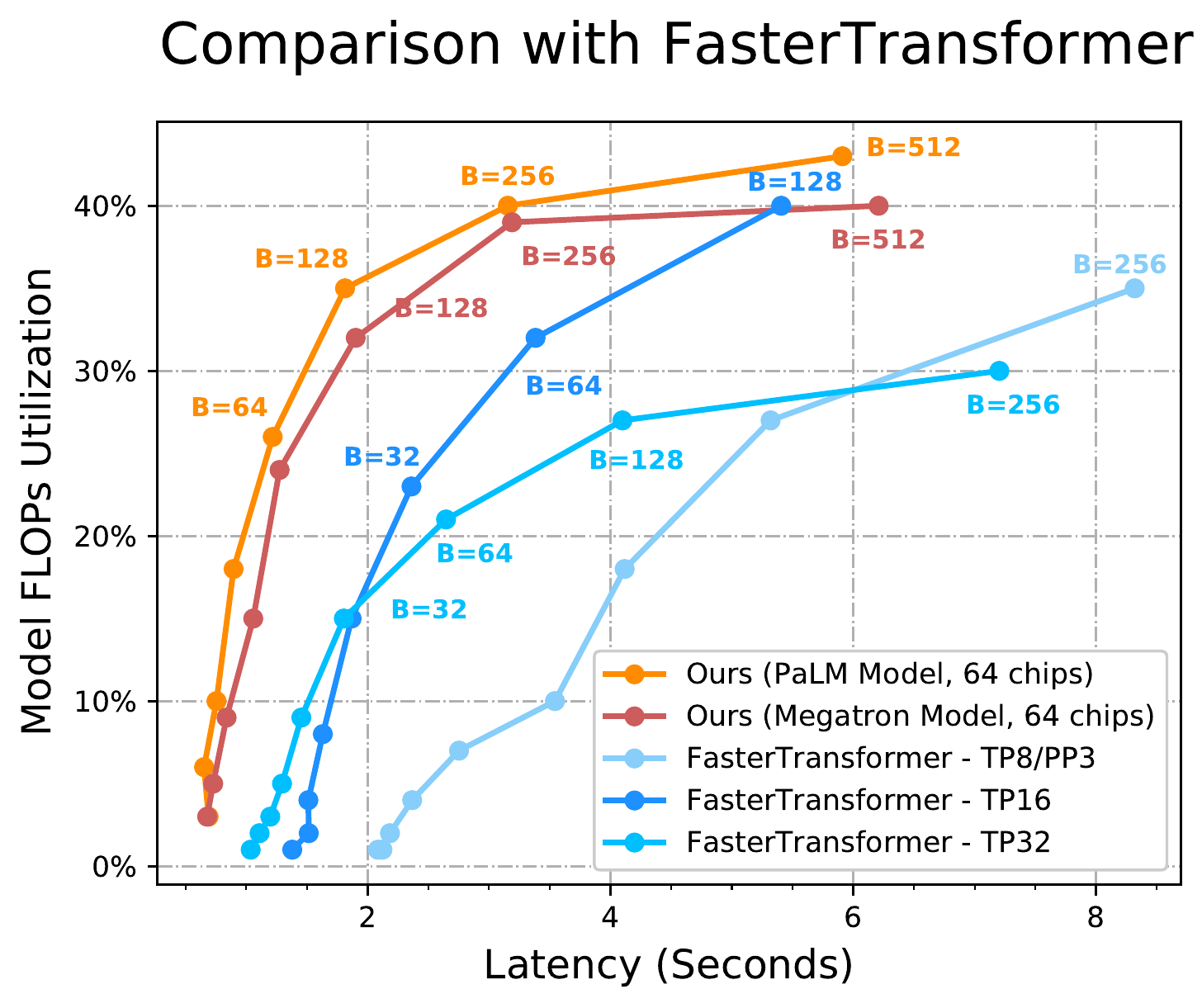}
\caption{Model FLOPS utilization (MFU) versus total latency for running a 60-input-token, 20-output-token inference, at a range of batch sizes.}
\label{fig:ours-vs-fastertransformer-60-20}
\vspace{-0.27in}
\end{figure}

Figure~\ref{fig:ours-vs-fastertransformer-60-20} shows the performance of our implementation relative to three FasterTransformer configurations. 
We benchmark the Megatron 530B model~\cite{smith2022using} and the similarly-sized PaLM 540B model, which has architectural optimizations including multiquery attention and parallel attention/feedforward layers (full list of differences in Table~\ref{table:hyperparams-vs-megatron}). Our implementation of PaLM 540B achieves the best absolute latency, and our implementation also offers the best MFU for the Megatron model for all but one latency target. Our PaLM implementation outperforms our Megatron implementation by up to 10\% MFU in this benchmark primarily because of the parallel attention/ffn layers. Compared to Section~\ref{sec:attention-ablation}, the advantage of parallel layers is partially offset by Megatron's larger $\dmodel$ and $\dff$ sizes. The advantage of multiquery attention is not noticeable in this benchmark because the attention context length is too short.

FasterTransformer reports results with 8-, 16-, and 32-way tensor parallelism. Their 32-way tensor parallelism achieves a maximum of 33\% MFU across all reported benchmarks, compared to 46\% MFU in their 16-way tensor parallel configuration. This likely indicates a communication bottleneck of scaling tensor parallelism beyond this point. In contrast, our implementation is able to scale up to 64-way tensor parallelism while still achieving 44\% MFU, suggesting superior scalability of our 2D weight-stationary partitioning strategy on TPU v4's larger high-speed interconnect domains.

We provide results on all the configurations used in the FasterTransformer baseline in Appendix~\ref{sec:appendix-fastertransformer}. We also note that our benchmarks throughout the paper attempt to include more challenging inference scenarios, such as context lengths in the range 1024--4096, and report the inference latency for the generate phase and the prefill phase separately (since they have different characteristics).

\section{Related Work}\label{sec:relatedwork}
\paragraph{Parallelism approaches.} Prior works propose several approaches for efficient partitioning to train large models efficiently, for e.g., NeMo Megatron~\cite{korthikanti2022reducing}, GSPMD~\cite{xu2021gspmd} and Alpa~\cite{zheng2022alpa}. \citet{fastertransformer} establishes a benchmark suite for multi-GPU multi-node inference for a range of different model sizes, including Megatron--Turing NLG 530B. The key inference speedups come from combining tensor parallelism and pipeline parallelism in conjuction with memory optimizations. DeepSpeed Inference \cite{aminabadi2022deepspeed} further enables ZeRO offload
to use CPU and NVMe memory in addition to the GPU memory. For larger batch sizes, \citet{effectivetransformer} packs consecutive sequences together to minimize padding. \citet{zheng2022alpa} generalizes the search through parallelism strategies via integer-linear programming. In comparison, this paper derives the partitioning strategies based on intuitive empirically-backed analytical tradeoffs  to meet the application requirements that scale well with model size, context length and chip count. %

\paragraph{ML inference efficiency.} Several approaches~\cite{gupta2020compression} to improve the inference 
efficiency of Transformer models focus on model architecture improvements, for example efficient attention layers~\cite{roy2020efficient,choromanski2020rethinking, kitaev2020reformer, sukhbaatar2019adaptive, child2019generating}, distillation~\cite{sanh2019distilbert,sun2020mobilebert}, and model compression techniques, such as pruning~\cite{li2020train, brix2020successfully,zhou2021learning, li2020efficient, wang2020hat}, or quantization~\cite{dettmers2022llm,abdolrashidi2021pareto,zafrir2019q8bert,zhang2018lq}. 
This paper reuses the prior work on model quantization to add to the inference speedups, and the techniques we describe could also be coupled with other model compression methods.

\section{Conclusions}\label{sec:conclusions}
\vspace{-0.05in}
Large Transformer-based models are unlocking new capabilities and applications in several domains, but we need significant advances to democratize their access as we scale up the model size. This paper investigates the scaling properties of Transformer inference workloads and proposes practical partitioning approaches to meet challenging application requirements such as tight latency targets (on the order of seconds for 500B+ parameter models). We show that the best latencies are achieved by going far beyond the traditional paradigm of single-server inference, and scaling inference up to 64+ chips. 
Longer context lengths incur higher memory costs, but multiquery attention with appropriate partitioning reduces this cost and makes long-context inference practical. 
The proposed partitioning strategies generalize to many topologies, including single- and multi-node NVLink networks in GPU systems. 

Although we achieve our goal of pushing the boundaries of scale for inference workloads, we observe that FLOP count and communication volume can fundamentally limit inference performance of dense Transformer models. Sparsity techniques, such as task-based mixture of expert architectures~\cite{fedus2022review, kudugunta2021beyond, lepikhin2020gshard, shazeer2017outrageously}, and adaptive computation techniques that allocate different amounts of compute per input and generation timestep~\cite{jaszczur2021sparse,schuster2022confident}, promise to reduce FLOPs per token of Transformer models. We are hopeful that such techniques that reduce FLOPs per token, as well as techniques that compress chip-to-chip communication, will enable further gains in both cost and latency.

\section{Acknowledgments}
Our work builds on top of the work of many, many teams at Google.  We'd especially like to recognize the PaLM team, T5X team, the Pathways infrastructure team, the JAX team, the Flaxformer team, the XLA team, and the AQT team. We are grateful to Blake Hechtman, Marcello Maggioni, Zongwei Zhou, and Shibo Wang for XLA support and performance optimizations. We would like to thank our colleagues for valuable inputs and discussion on the project -- Jacob Austin, Yuanzhong Xu, Lukasz Lew, Sharan Narang, Adam Roberts, Noah Fiedel, and Mike Gunter. We also thank Hyeontaek Lim, James Laudon, George Necula, Martin Abadi and Chandu Thekkath for their review and feedback in improving the presentation of this work, and Erica Moreira for the support of compute resources.

\pagebreak
\bibliographystyle{plainnat}
\bibliography{references}
\clearpage
\vfill
\pagebreak
\newpage
\appendix
\appendix
\renewcommand\thefigure{\thesection.\arabic{figure}}
\renewcommand\thetable{\thesection.\arabic{table}}
\section{Partitioning Strategies: Deriving Communication Costs}\label{sec:appendix-partitioning}
\setcounter{figure}{0}
\setcounter{table}{0}
\subsection{Cost of all-gather/reduce-scatter}
\label{sec:appendix-cost-allreduce}

\begin{figure*}[t]
\centering
\includegraphics[height=2in]{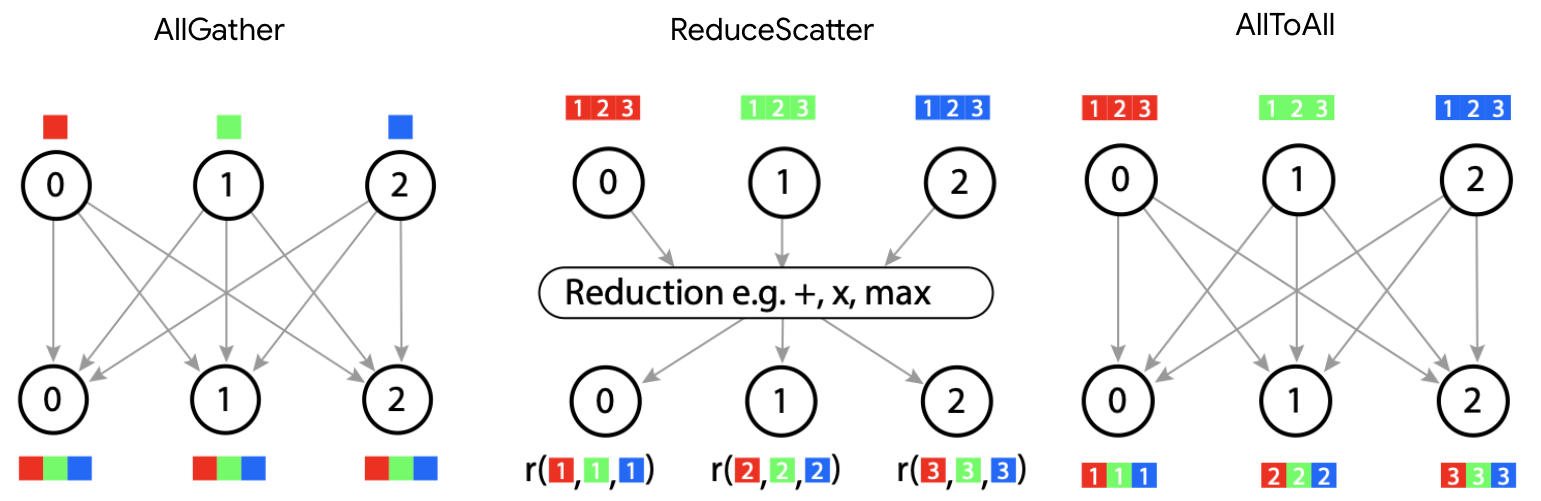}
\caption{Communication patterns of collective operations: all-gather, reduce-scatter, and all-to-all across three devices.}
\label{fig:collectives}
\vspace{-0.2in}
\end{figure*}
Figure~\ref{fig:collectives} shows typical collective operations we use in partitioning strategies and their communication patterns across three devices. For an all-gather over $K$ partitions, where each chip produces an output of size $D$, 
the communication pattern requires chunks of size $\frac{D}{K}$ to be transferred over $(K-1)$ interconnect links in the process of getting copied to $(K-1)$ chips.
The resulting communication time for the all-gather is
\begin{align*}
    T_\textrm{comm(all-gather)} &= \dfrac{D}{ (\textrm{network bandwidth}) }\dfrac{K-1}{K}.
\end{align*}    
This is a general cost model that holds true for most real-world network topologies~\cite{chan2007collective}, not just the TPU's torus topology.

The communication time for a reduce-scatter $T_\textrm{comm(reduce-scatter)}$ is the same, except that $D$ is the size of the (larger) input buffer rather than the (smaller) output buffer. Thus, the total communication time for an all-reduce is $T_\textrm{comm(all-reduce)} = 2 \times T_\textrm{comm(all-gather)}$. 

In most formulas, we will disregard the $(K-1)/K$ term, approximating it as 1 under the assumption $K\gg 1$, in order to simplify the algebra. This yields a simple approximation: reduce-scatter time is proportional to the size of the per-chip input, and all-gather time is proportional to the size of the per-chip output.
\subsection{Details for communication time calculations}
\subsubsection{Feedforward layer, 2D weight-stationary layout}
\label{sec:appendix-partitioning-ws2d}
Figure~\ref{fig:ffn}(b) shows the partitioning layout. The partitioning layout for weights is $E_{x}F_{yz}$, 
i.e. they are partitioned along the $\dmodel$ dimension into $X$ partitions and 
along the $\dff$ dimension into $Y\times Z$ partitions, 
where $X \times Y\times Z = \nchips$. We now show how to size the $X$, $Y$ and $Z$ axes of the torus to minimize total communication time in 2D weight-stationary layout. The communication time is:
\begin{align*}
    T_\textrm{comm} &= \frac{2BL}{\textrm{network bandwidth}}\left(\frac{E}{X}+\frac{F}{YZ}\right)
\end{align*}

We have a free choice of $X$, $Y$ and $Z$ subject to available TPU v4 slice shapes and $X \times Y \times Z = \nchips$.
Assuming $\dff=4 \times \dmodel$, 
we achieve the minimum communication time with $X=0.5\times\sqrt{\nchips}$ 
and $YZ=2\times\sqrt{\nchips}$. The resulting total communication time is:
\begin{align*}
    T_\textrm{comm} &= \dfrac{8 B L E}{\sqrt{\nchips}\times \textrm{network bandwidth}}.
\end{align*}

\subsubsection{Feedforward layer, weight-gathered layout}
\label{sec:appendix-weight-gathered}
\begin{figure}[H]
\centering
\includegraphics[height=2.4in]{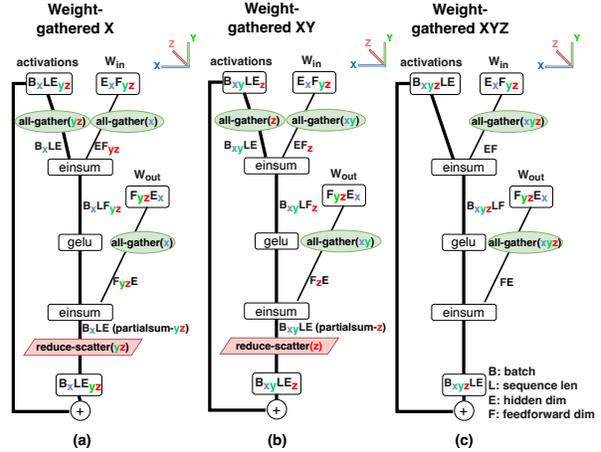}
\caption{Weight-gathered layouts for Feedforward layer.}
\label{fig:ffnweightgathered}
\vspace{-0.1in}
\end{figure}

Figure~\ref{fig:ffnweightgathered} shows the different weight-gathered layouts, while Figure~\ref{fig:ffn}(c) shows one instance of XY weight-gathered layout. A key aspect of the specific layout we choose is that weights start in the same $E_xF_{yz}$ layout as in 2D weight-stationary, so that we can instantly switch between weight-gathered layout and weight-stationary layout. Just before the einsums, the weight tensors are all-gathered over the $X$ and $Y$ axes, with communication volume $EF/Z$.  

By changing the relative sizes of the $X$, $Y$, and $Z$ axes, we can trade off weight communication against activation communication, and thereby minimize the total communication volume. We now show the asymptotic scaling of weight-gathered layouts. Let $N$ be the number of chips that weights are all-gathered over: $N=X$ in $X$-weight-gathered, $N=XY$ in $XY$-weight-gathered, and $N=XYZ$ in $XYZ$-weight-gathered. 

Weight communication is:
\begin{align*}
T_\textrm{comm(weights)} & = \dfrac{2EF\times N}{\nchips\times \textrm{network bandwidth}}.
\end{align*}
Activation communication is:
\begin{align*}
T_\textrm{comm(acts)} & = \dfrac{2 B L E}{N\times \textrm{network bandwidth}}.
\end{align*}

Total communication is minimized by the choice $N=\sqrt{BS\nchips/F}$, which yields total communication time
\begin{align*}
T_\textrm{comm} & = 4 E \dfrac{\sqrt{B L F}}{\sqrt{\nchips}\times \textrm{network bandwidth}}
\end{align*}

Figure~\ref{fig:communication-volume-by-batch} shows how the communication-optimal configuration switches between these layouts as batch size grows. While the 2D weight-stationary strategy minimizes communication at low tokens per batch, different weight-gathered layouts are optimal at larger number of tokens per batch.
\newpage
\section{Minimum prefill latency}\label{sec:appendix-minimum-prefill-latency}
\setcounter{figure}{0}
\setcounter{table}{0}
We report here the minimum latency required for prefill. Figure~\ref{fig:prefill-cost-vs-latency-batch-1} shows the Pareto frontier of cost vs. latency as we sweep sequence length from 32 to 1024 at batch size 1.
\begin{figure}[H]
\centering
\includegraphics[height=2in]{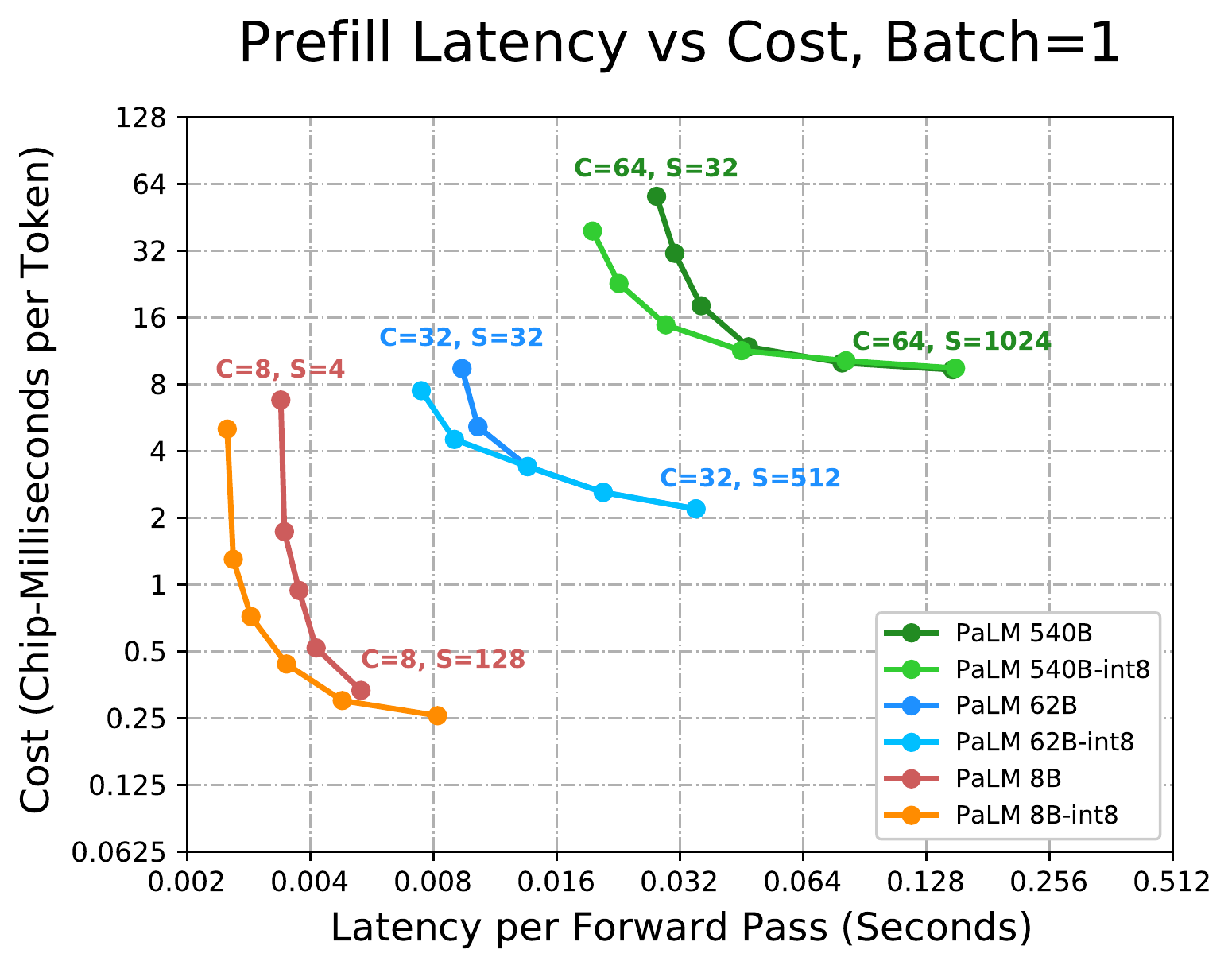}
\caption{Prefill cost vs. latency for PaLM models over a range of sequence lengths S. C indicates chip count.}
\label{fig:prefill-cost-vs-latency-batch-1}
\end{figure}

\section{MFU vs latency tradeoff}\label{sec:appendix-mfu-vs-latency}
\setcounter{figure}{0}
\setcounter{table}{0}
\begin{figure*}[tbh]
\centering
    \begin{subfigure}{0.49\linewidth}
  \centering
  \includegraphics[height=2in]{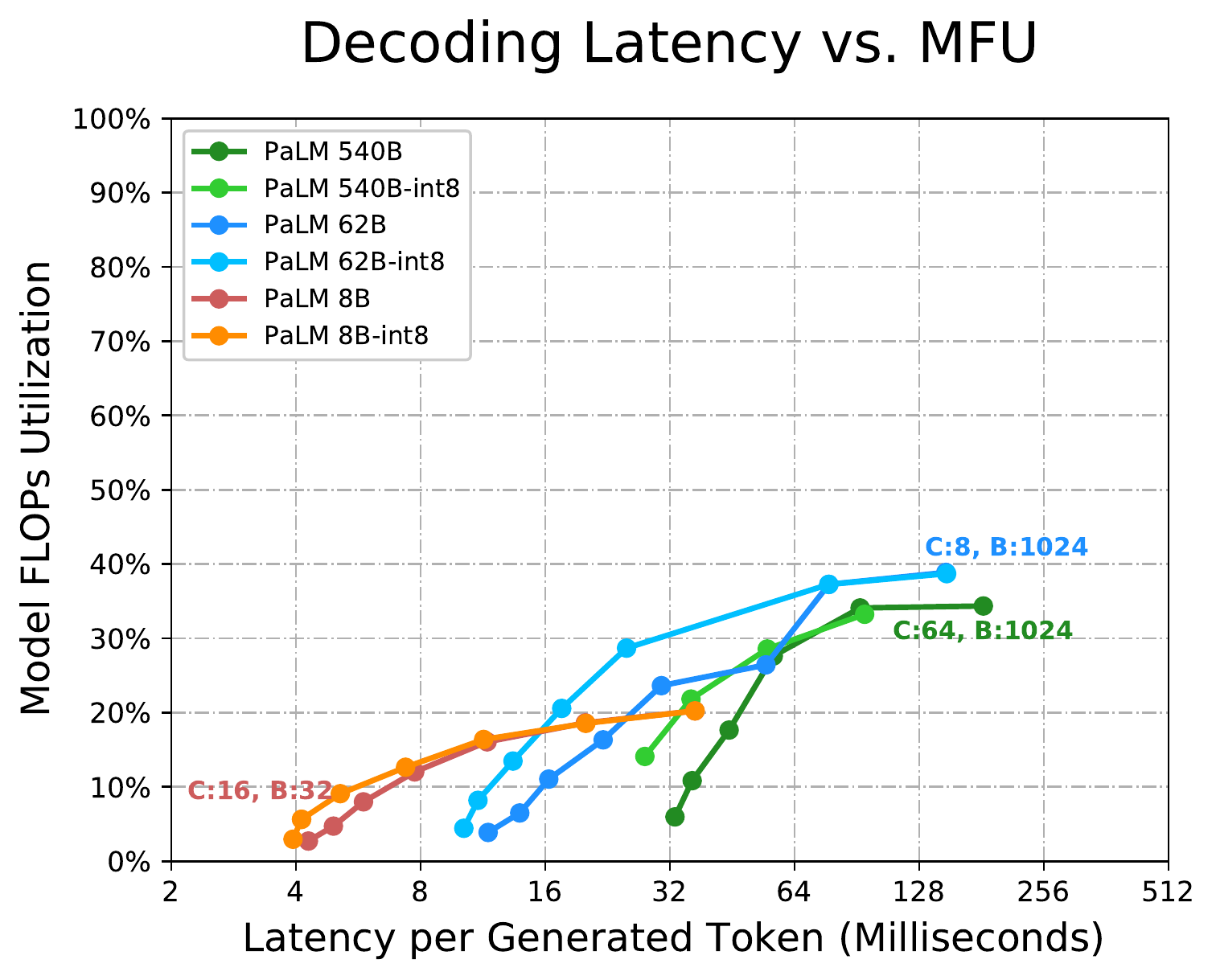}
  \label{fig:generate-mfu-vs-latency}
  \end{subfigure}
\begin{subfigure}{0.49\linewidth}
  \centering
  \includegraphics[height=2in]{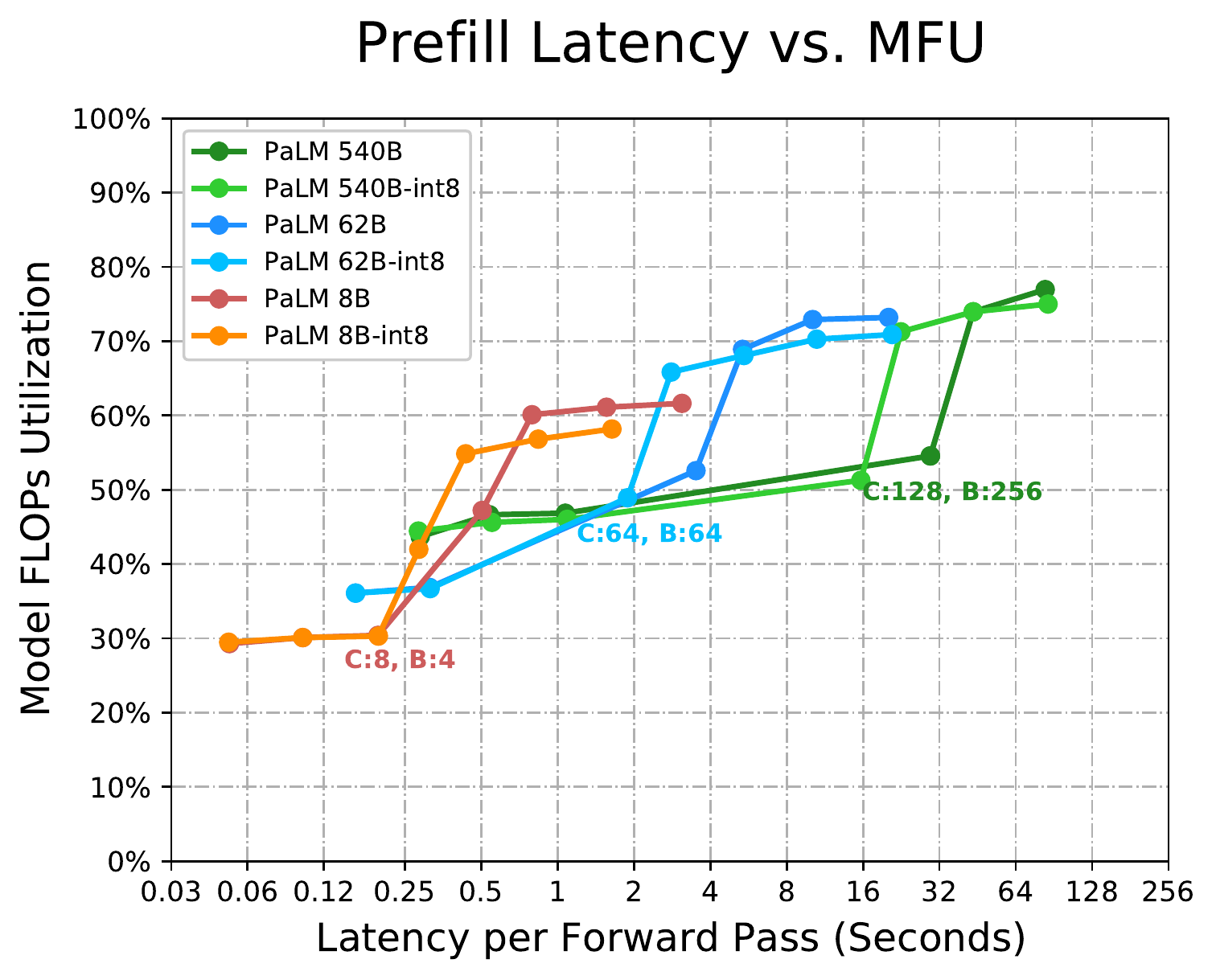}
  \label{fig:prefill-mfu-vs-latency}
  \end{subfigure}
  \caption{MFU vs.\ latency for PaLM models. We use a context length of 2048. Points in each line represent the Pareto frontier of efficiency versus latency. Chip count is $C$, batch size is $B$. Left: latency per token for generating 64 tokens, assuming the context has already been processed. Right: time to process 2048 input tokens; excludes the time to generate any output tokens. The corresponding cost vs latency numbers are shown in Figure~\ref{fig:cost-vs-latency}. }
  \label{fig:mfu-vs-latency}
\end{figure*}
We report here the relationship between model size, latency, and MFU. Figure~\ref{fig:mfu-vs-latency} shows the Pareto frontier of MFU vs. latency as we sweep the batch size and the number of chips same as Figure~\ref{fig:cost-vs-latency}. The MFU for decode  is typically much lower than for prefill. In the prefill phase, the ``jumps'' in MFU show the transition point from weight stationary 2D layout to XYZ weight gathered layout.

In most cases, the larger models achieve higher MFUs than the smaller models, because larger matrix multiplies are more efficient. However, at long-latency decodes, PaLM 62B achieves higher MFU than PaLM 540B, because the former uses 8-way model parallelism and the latter uses 64-way model parallelism. We may be able to further optimize PaLM 540B by reducing the model parallelism in the high-throughput (latency-tolerant) regime.

\newpage
\section{Full comparison to FasterTransformer}\label{sec:appendix-fastertransformer}
\setcounter{figure}{0}
\setcounter{table}{0}
 In this section, we report the latency and MFU of our implementations of both the PaLM~540B model and the Megatron-Turing NLG 530B model run on 64 TPU v4 chips, in comparison to FasterTransformer baselines. We first note the model architecture differences in Table~\ref{table:hyperparams-vs-megatron}. 
 
Then, we report the the full set of comparisons for the three configurations in the FasterTransformer benchmarks: 20 input tokens and 8 output tokens in  Table~\ref{table:fastertransformer-comparison-20-vs-8}, 60 input tokens and 20 output tokens in  Table~\ref{table:fastertransformer-comparison-60-vs-20}, and 128 input tokens and 8 output tokens in Table~\ref{table:fastertransformer-comparison-128-vs-8}.

For each table we report the \emph{Pareto frontier} of latency and MFU with \textbf{bold font} (frontier across all 500B-class results) and \underline{underline} (frontier across MT-NLG specifically). This frontier is \emph{not} a per-row comparison, but instead is defined globally across the table. It is defined as follows: a benchmark result $(\mathrm{latency}, \mathrm{MFU})$ is on the Pareto frontier if, for all other benchmark results $(\mathrm{latency}_2,\mathrm{MFU}_2)$, either $\mathrm{latency}\leq \mathrm{latency}_2$ or $\mathrm{MFU}\geq \mathrm{MFU}_2$ (or both) is true. Visually, this corresponds to being ``up and to the left'' in Figure~\ref{fig:ours-vs-fastertransformer-60-20}.

We do not report batch sizes below 4 because our partitioning strategy partitions multiquery attention over batch and achieves no speedup for a batch size smaller than 4 (the minimum size of a TPU v4 torus axis).

\begin{table}[hb!]
\centering
\footnotesize
\begin{tabular}{l | p{0.9in} p{1.2in}}
 & \textbf{PaLM 540B} & \textbf{Megatron 530B} \\
\hline
$\nparams$ & 540B & 530B \\
$\nlayers$ & 118 & 105 \\
$\dmodel$ & 18432 & 20480 \\
$\dff$ & 73728 & 81920 \\
$\nheads$ & 48 & 128 \\
$\dhead$ & 256 & 160 \\
\hline
Attention & Multiquery & Multihead \\
Parallel ffn/attn & Yes & No \\
\end{tabular}
\caption{Hyperparameters for PaLM and Megatron-Turing NLG inference.}
\label{table:hyperparams-vs-megatron}
\end{table}

\begin{table*}[t]
\small
\begin{tabular}{r|rrrrrr|rrrrrrrr}
 & \multicolumn{6}{c|}{FasterTransformer MT-NLG 530B total} & \multicolumn{8}{c}{Ours (530B/540B on 64 TPU v4 with 2D partitioning)} \\
 & \multicolumn{2}{c}{TP16} & \multicolumn{2}{c}{TP32} & \multicolumn{2}{c|}{PP3/TP8} & \multicolumn{2}{c}{PaLM prefill} & \multicolumn{2}{c}{PaLM generate} & \multicolumn{2}{c}{PaLM total} & \multicolumn{2}{c}{MT-NLG total} \\
batch & time & MFU & time & MFU & time & MFU & time & MFU & time & MFU & time & MFU & time & MFU\\ 
\hline
1 & 	565 & 	1\% & 	431 & 	1\% & 	842 &	0\% &	- & 	-	& - & 	- & 	-	 & - & -	 & - \\
2 & 	598 & 	2\% & 	455 & 	1\% & 	860 &	1\% &	- & 	-	& - & 	- & 	-	 & - & -	 & - \\
4 & 	616 & 	4\% & 	493 & 	2\% & 	867 &	2\% &	34 & 	14\% & 	255 & 	1\% & 	\textbf{289} &	\textbf{2\%}  & \underline{\textbf{289}} &	 \underline{\textbf{2\%}}\\
8 & 	660 & 	7\% & 	523 & 	5\% & 	929 &	3\% &	40 & 	25\% & 	226 & 	2\% & 	\textbf{265} &	\textbf{5\%}  & \underline{304} &	 \underline{4\%}\\
16 & 	730 & 	13\% & 	575 & 	8\% & 	1049 &	6\% &	58 & 	34\% & 	234 & 	3\% & 	\textbf{292} &	\textbf{9\%}  & \underline{339} &	 \underline{8\%}\\
32 & 	865 & 	22\% & 	672 & 	14\% & 	1283 &	10\% &	99 & 	40\% & 	235 & 	7\% & 	\textbf{334} &	\textbf{16\%}  & \underline{420} &	 \underline{13\%}\\
64 & 	1191 & 	32\% & 	942 & 	20\% & 	1722 &	15\% &	186 & 	42\% & 	265 & 	12\% & 	\textbf{451} &	\textbf{24\%}  & \underline{532} &	 \underline{20\%}\\
128 & 	\underline{1862} & 	\underline{41\%} & 	1431 & 	27\% & 	2124 &	24\% &	356 & 	44\% & 	312 & 	20\% & 	\textbf{668} &	\textbf{33\%}  & \underline{740} &	 \underline{29\%}\\
256 & 	\underline{\textbf{3341}} & 	\underline{\textbf{46\%}} & 	2483 & 	31\% & 	3140 &	32\% &	668 & 	47\% & 	415 & 	30\% & 	\textbf{1083} &	\textbf{41\%}  & \underline{1151} & 	\underline{38\%}\\
512 & 	- & 	- &	- & 	- & 	-	 & - &	1366	& 46\%	& 671 & 	37\% 	& \textbf{2037} &	\textbf{43\%}  & 2151 & 	40\%\\
1024 & 	- & 	- &	- & 	- & 	-	 & - &	2785	& 45\%	& 1257 & 	40\% 	& 4041 &	44\%  & 4082 &	 42\%\\
\end{tabular}
\caption{Results for the 20-input-token, 8-output-token benchmark. All times in milliseconds. The \textbf{bold} and \underline{underline} annotations are \emph{not} per row, but instead show the Pareto frontier of time vs.\ MFU. See Section~\ref{sec:appendix-fastertransformer} for full explanation.}
\label{table:fastertransformer-comparison-20-vs-8}
\end{table*}

\begin{table*}[t]
\footnotesize
\begin{tabular}{r|rrrrrr|rrrrrrrr}
 & \multicolumn{6}{c|}{FasterTransformer MT-NLG 530B total} & \multicolumn{8}{c}{Ours (530B/540B on 64 TPU v4 with 2D partitioning)} \\
 & \multicolumn{2}{c}{TP16} & \multicolumn{2}{c}{TP32} & \multicolumn{2}{c|}{PP3/TP8} & \multicolumn{2}{c}{PaLM prefill} & \multicolumn{2}{c}{PaLM generate} & \multicolumn{2}{c}{PaLM total} & \multicolumn{2}{c}{MT-NLG total} \\
batch & time & MFU & time & MFU & time & MFU & time & MFU & time & MFU & time & MFU & time & MFU\\ 
\hline
1 & 	1379 & 	1\% & 	1037 & 	1\% & 	2085 &	1\% &	- & 	- &	- & 	- & 	-	 & - & -	 & - \\
2 & 	1515 & 	2\% & 	1110 & 	2\% & 	2122 &	1\% &	- & 	-	& - & 	- & 	-	 & - & -	 & - \\
4 & 	1512 & 	4\% & 	1198 & 	3\% & 	2184 &	2\% &	50 & 	29\% & 	640 & 	1\% & 	690 &	3\% & \underline{678}	& \underline{3\%} \\
8 & 	1631 & 	8\% & 	1295 & 	5\% & 	2367 &	4\% &	80 & 	37\% & 	574 & 	2\% & 	\textbf{653} &	\textbf{6\%} & \underline{728}	& \underline{5\%} \\
16 & 	1868 & 	15\% & 	1454 & 	9\% & 	2753 &	7\% &	153 & 	39\% & 	602 & 	3\% & 	\textbf{755} &	\textbf{10\%} & \underline{838}	& \underline{9\%} \\
32 & 	2361 & 	23\% & 	1804 & 	15\% & 	3543 &	10\% &	270 & 	44\% & 	626 & 	6\% & 	\textbf{896} &	\textbf{18\%} & \underline{1058}	& \underline{15\%} \\
64 & 	3383 & 	32\% & 	2646 & 	21\% & 	4117 &	18\% &	501 & 	47\% & 	717 & 	11\% & 	\textbf{1218} &	\textbf{26\%} & \underline{1275}	& \underline{24\%} \\
128 & 	\underline{5406} & 	\underline{40\%} & 	4099 & 	27\% & 	5319 &	27\% &	985 & 	48\% & 	829 & 	19\% & 	\textbf{1814} &	\textbf{35\%} & \underline{1902}	& \underline{32\%} \\
256 & 	OOM & 	- &	7203 &	30\% & 	8318\	 & 35\% & 2041 & 	46\% & 	1114 & 	28\% & 	\textbf{3155} &	\textbf{40\%} & \underline{3189}	& \underline{39\%} \\
512 & 	- & 	- &	- & 	- & 	-	 & - &	4167 &	 45\% &	1743 & 	36\% &	 \textbf{5910} &	\textbf{43\%} & 6210	& 40\% \\
1024 & 	- & 	- &	- & 	- & 	-	 & - &	8349 &	 45\% &	3260 & 	39\% &	 11608 &	43\% & 12390	& 40\% \\
\end{tabular}
\caption{Results for the 60-input-token, 20-output-token benchmark. All times in milliseconds. The \textbf{bold} and \underline{underline} annotations are \emph{not} per row, but instead show the Pareto frontier of time vs.\ MFU. See Section~\ref{sec:appendix-fastertransformer} for full explanation.}
\label{table:fastertransformer-comparison-60-vs-20}
\end{table*}

\begin{table*}[t]
\small
\begin{tabular}{r|rrrrrr|rrrrrrrr}
 & \multicolumn{6}{c|}{FasterTransformer MT-NLG 530B total} & \multicolumn{8}{c}{Ours (530B/540B on 64 TPU v4 with 2D partitioning)} \\
 & \multicolumn{2}{c}{TP16} & \multicolumn{2}{c}{TP32} & \multicolumn{2}{c|}{PP3/TP8} & \multicolumn{2}{c}{PaLM prefill} & \multicolumn{2}{c}{PaLM generate} & \multicolumn{2}{c}{PaLM total} & \multicolumn{2}{c}{MT-NLG total} \\
batch & time & MFU & time & MFU & time & MFU & time & MFU & time & MFU & time & MFU & time & MFU\\ 
\hline
1 & 	585 & 	5\% & 	451 & 	3\% & 	866 &	2\% &	- & 	-	& - & 	- & 	-	 & - \\
2 & 	667 & 	9\% & 	508 & 	6\% & 	932 &	4\% &	- & 	- &	- & 	- & 	-	 & - \\
4 & 	765 & 	15\% & 	606 & 	10\% & 	1097 &	7\% &	81 & 	39\% & 	258 & 	1\% & 343	& 10\% & 	\textbf{\underline{338}} &	\textbf{\underline{10\%}} \\
8 & 	990 & 	23\% & 	766 & 	15\% & 	1434 &	11\% &	149 & 	42\% & 	234 & 	2\% & 403	& 17\% & 	\textbf{\underline{384}} &	\textbf{\underline{16\%}} \\
16 & 	1377 & 	34\% & 	1074 & 	22\% & 	2104 &	15\% &	287 & 	44\% & 	253 & 	3\% & 586	& 23\% & 	\textbf{\underline{540}} &	\textbf{\underline{23\%}} \\
32 & 	2251 & 	41\% & 	1741 & 	27\% & 	2623 &	23\% &	536 & 	47\% & 	263 & 	6\% & \textbf{796}	& \textbf{34\%} & 	\underline{799} &	\underline{33\%} \\
64 & 	\underline{4002} & 	\underline{46\%} & 	3114 & 	30\% & 	3578 &	34\% &	1056 & 	48\% & 	317 & 	10\% & \textbf{1329}	& \textbf{40\%} & 	\underline{1372} &	\underline{39\%} \\
128 & 	OOM & 	- & 	5784 & 	32\%	& 5512	& 45\%	& 2202	& 46\%	 & 381	 & 17\%	 & \textbf{2343}	& \textbf{46\%} & \underline{2583}	& \underline{45\%} \\
256 & 	OOM & 	- & 	11232 & 	33\%	& \textbf{\underline{9614}} &	\textbf{\underline{51\%}} &	4479	& 45\%	 & 431	 & 29\%	 & 4710	& 45\% & 4911	& 45\% \\
512 & 	- & 	- & 	- & 	-	& -	& -	& 8913	& 45\%	 & 734	 & 34\%	 & 9673	& 44\% & 9647	& 43\% \\
1024 & 	- & 	- & 	- & 	-	& -	& -	& 17766	& 45\%	 & 1370	 & 37\%	 & 19723	& 43\% & 19136	& 43\% \\
\end{tabular}
\caption{Results for the 128-input-token, 8-output-token benchmark. All times in milliseconds. The \textbf{bold} and \underline{underline} annotations are \emph{not} per row, but instead show the Pareto frontier of time vs.\ MFU. See Section~\ref{sec:appendix-fastertransformer} for full explanation.}
\label{table:fastertransformer-comparison-128-vs-8}
\end{table*}

\end{document}